\documentclass{article} 
\usepackage{iclr2026_conference,times}

\usepackage{amsmath,amsfonts,bm}

\def\eqref#1{equation~\ref{#1}}

\def\1{\bm{1}}

\DeclareMathAlphabet{\mathsfit}{\encodingdefault}{\sfdefault}{m}{sl}
\SetMathAlphabet{\mathsfit}{bold}{\encodingdefault}{\sfdefault}{bx}{n}

\usepackage{hyperref}
\usepackage{url}
\usepackage{graphicx}
\usepackage{booktabs}
\usepackage{amsmath}
\usepackage{enumitem}
\usepackage{multirow}
\usepackage{wrapfig}
\usepackage[most]{tcolorbox}

\newtcolorbox{findingbox}{
  colback=gray!1,
  colframe=blue!10!black,
  boxrule=0.5pt,
  arc=1.5pt,
  left=6pt,
  right=6pt,
  top=5pt,
  bottom=5pt,
  before skip=6pt,
  after skip=8pt
}
\title{Do Physics Foundation Models Learn Generalizable Physics?\\
A Bias-Aware Benchmark Across Physical Regimes and Distribution Shifts
}

\author{
Mengdi Chu$^{1}$ \quad
Yang Liu$^{1}$ \quad
Ayan Biswas$^{2}$ \quad
Han-Wei Shen$^{1}$ \\
$^{1}$The Ohio State University \quad
$^{2}$Los Alamos National Laboratory \\
\texttt{\{chu.752, liu.12870, shen.94\}@osu.edu} 
\texttt{ayan@lanl.gov}
}

\iclrfinalcopy
\begin{document}

\maketitle

\begin{abstract}
Recent physics foundation models claim general spatiotemporal forecasting ability, yet their evaluations often collapse performance into a single average score under a fixed training distribution. This makes it difficult to determine whether a model has learned generalizable physical dynamics or only performs well under particular settings.
We construct a benchmark with 8 physical dynamics, 3 training-data mixtures, 
and 25 test regimes induced by dynamic-scale and initial-condition complexity shifts, 
covering in-distribution, distribution-shift, and out-of-distribution settings. 
We evaluate five physics foundation model architectures and four model variants per architecture (scratch and three pretrained sizes), resulting in $60{,}000$ 
measurements.
Our results show that current physics foundation models behave as conditional rather than universal generalists: their generality depends on the physical regime, temporal scale, initial-condition setting, pretraining, model size, and architecture. 
Improving the training data distribution only partially mitigates this limitation. Pretraining and scaling are also unable to reliably remove their ability biases.
We argue that improving physics foundation models requires moving beyond scaling models or expanding data, toward learning mechanisms that better capture transferable physical knowledge across regimes, temporal scales, and distribution shifts.
\textbf{PhysBiasBench} can be accessed at
\url{https://huggingface.co/datasets/90879c/PhysBiasBench}.


\end{abstract}

\section{Introduction}\label{sec:intro}

Foundation models are increasingly being extended from language and vision to
physical forecasting and scientific simulation.
Across learned physical simulators
\citep{battaglia2016interaction,sanchezgonzalez2020learning,pfaff2021meshgraphnets},
data-driven physical forecasting systems
\citep{pathak2022fourcastnet,lam2023graphcast,bodnar2025aurora},
neural operators and neural PDE solvers
\citep{lu2021deeponet,li2021fno,kovachki2023neuraloperator,brandstetter2022message},
and PDE foundation models
\citep{hao2024dpot,herde2024poseidon,mccabe2023multiple}, these models are often motivated by the hope that one trained model can be reused across physical regimes rather than rebuilt for each equation or setting. 
This motivates a central
question:  \emph{do physics foundation models learn transferable physical knowledge, or do they mainly learn strong predictors for the distributions they have seen?}
However, current evaluation protocols often make this question difficult to answer. Existing PDE benchmarks broaden the set of equations and tasks
\citep{takamoto2022pdebench,thewell2024,koehler2024apebench,hassan2023bubbleml},
but model comparisons are often reduced to a single average score under one training distribution.
Such averages are useful for leaderboard ranking, but they
hide the conditional structure of physical generalization. 
For example, a model may perform well on some physical regimes while failing on others; 
it can perform well at a familiar temporal scale but fail under temporal extrapolation; 
it can benefit from pretraining in train-seen regimes while transferring harmful biases under distribution shift.
Understanding these capability patterns and generalization boundaries is especially important in physical systems.

We therefore study \emph{conditional generalization} in physics foundation models and separate three questions that are usually
coupled. First, \textit{is model capability uniformly general, or does it exhibit systematic capability biases across PDE families and distribution-shift types?} Second, \textit{can changing the training-data distribution close the out-of-distribution gap, or does it mainly improve familiar regimes?} Third, \textit{do model interventions such as pretraining and scaling fix these biases, or do they introduce their own conditional preferences?}
Rather than asking only which model has the lowest error, we ask where each model generalizes, where it fails, and which factors explain the failure.

We design the benchmark to vary the main axes along which generality can break. We generate the PDE trajectories with the APEBench/Exponax toolkit~\citep{koehler2024apebench}. The physical axis contains 8 PDE dynamics spanning reaction--diffusion, fluid-like, chaotic, and wave-like regimes. The data axis contains three training mixtures with different biases toward simple, balanced, or complex regimes. The test axis forms a $5\times5$ grid over dynamic-scale and initial-condition complexity, covering train-seen cells, interpolation-style distribution shifts, and extrapolative OOD regimes. 
The temporal axis evaluates both in-horizon prediction and longer-horizon rollout beyond the training rollout length.
On top of these data axes, we evaluate five physics foundation model
architectures: DPOT~\citep{hao2024dpot}, GPhyT~\citep{wiesner2025gphyt},
MORPH~\citep{rautela2025morph}, MPP~\citep{mccabe2023multiple}, and Poseidon~\citep{herde2024poseidon}, with four variants per architecture, including scratch models and three pretrained sizes from small to large. Together, this design yields $60{,}000$ aggregated measurements across physical dynamics, training mixtures, test regimes, model variants, and prediction horizons.

Our results show that current physics foundation models behave as conditional
rather than universal generalists. Their performance depends strongly on the
physical regime, temporal scale, initial-condition setting, training mixture,
pretraining state, model size, and architecture. 
For example, under the Mix-balance setting, within-model normalized error is consistently low on Fisher--KPP and Gray--Scott, but rises sharply on Wave dynamics, reaching
above $2\times$ the model-average error for several variants. 
Distribution shift further exposes this conditionality: dynamic-scale OOD regimes produce much larger ShiftDamage than initial-condition shifts, with the hardest cells reaching
roughly $7$--$8\times$ the train-seen reference error.
Increasing training
complexity only partially mitigates the problem; moving from Mix-simple to
Mix-complex reduces in-distribution raw error but increases normalized
ShiftDamage across OOD groups.
Model-side interventions are also conditional: under matched-size comparison,
$37.5\%$ of architecture--PDE pairs show negative pretraining transfer, and
larger pretrained variants are worse than the small baseline in $25.0\%$ of
larger-model cells.
These findings suggest that the limitation of current physics foundation
models is not only insufficient data or model size, but limited ability to learn
transferable physical structure across regimes, temporal scales, and distribution
shifts. This paper makes three contributions:

\section{Related Work}\label{sec:related}

The foundation-model paradigm is increasingly being extended from language and vision to scientific and physical domains, 
with the goal of moving from task-specific solvers toward reusable models across equations and regimes
\citep{bommasani2021opportunities}. 
Neural operators such as
DeepONet~\citep{lu2021deeponet} and the Fourier Neural Operator~\citep{li2021fno} learn maps between function spaces, while physics-informed and scientific-ML
methods use physical constraints to regularize learning \citep{raissi2019pinn,willard2022integrating}. Recent PDE foundation models
extend this direction by pretraining transformer- or operator-style architectures across multiple equations and regimes. DPOT~\citep{hao2024dpot},
Poseidon~\citep{herde2024poseidon}, GPhyT~\citep{wiesner2025gphyt}, MORPH~\citep{rautela2025morph}, and MPP~\citep{mccabe2023multiple} all aim at broad PDE
forecasting ability, but their evaluations are often summarized by average error under one training distribution.

An important ambiguity in learned physical models is whether low error reflects physical reasoning or high-quality pattern matching. 
Low forecasting error does not necessarily imply transferable physical understanding. Neural networks can exploit shortcuts, spurious correlations, or
distribution-specific regularities while still achieving high benchmark accuracy
\citep{shah2020pitfalls,geirhos2020shortcut}. This concern is important
for PDE forecasting, where a model may fit familiar regimes well but fail when the temporal scale, initial-condition
complexity, or rollout horizon changes. Existing evaluations often make this
hard to see because they report aggregate error under a limited set of training
and test conditions.

Scientific-ML benchmarks such as PDEBench~\citep{takamoto2022pdebench}, The Well~\citep{thewell2024}, APEBench~\citep{koehler2024apebench},
BubbleML~\citep{hassan2023bubbleml}, and CFDBench~\citep{cfdbench2023} have expanded the empirical coverage of learned physical models. 
A related line of work asks not only how accurate a model is, but what capability
patterns and biases are hidden by aggregate scores. Domain-generalization and
group-robustness benchmarks such as WILDS~\citep{koh2021wilds} and
DomainBed~\citep{gulrajani2020search} expose group-dependent failures under distribution shift, while model-bias studies show that high-level performance
can reflect shortcut cues or pretrained priors rather than input-grounded
reasoning~\citep{geirhos2020shortcut,shah2020pitfalls,vo2025vlmsbiased}. Recent physical-understanding benchmarks for vision-language models, such as
PhysBench~\citep{chow2025physbench} and
QuantiPhy~\citep{li2025quantiphy}, similarly evaluate whether general-purpose
models possess reliable physical capabilities rather than plausible responses.
However, existing physical foundation model evaluations rarely factorize the sources of capability. As a result, it
remains difficult to tell whether a model's apparent generality reflects
transferable physical capability or strong performance under particular regimes.
Therefore, our work connects these ideas in the setting of PDE foundation models. We evaluate apparent
generality through controlled changes in physical dynamics, training mixture,
test regime, prediction horizon, pretraining state, model size, and architecture.


\section{Benchmark and Protocol Design}\label{sec:bench}

We factorize model evaluation into physical dynamics, training distribution,
test-regime shift, prediction horizon, pretraining state, model size, and
architecture. This section defines these axes and the diagnostic metrics used in
the rest of the paper.

\subsection{Factorized evaluation axes}
Each measurement is indexed by
\[
E_H(m, v, p, r, c),
\]
where $m$ is the architecture, $v$ is the model variant, $p$ is the physical
dynamics, $r$ is the training mixture, and $c=(d,q)$ is the test cell. Here,
$d$ denotes the dynamic-scale setting and $q$ denotes initial-condition
complexity. $H$ denotes the prediction horizon.
~\autoref{tab:eval_axes} summarizes the role of each axis.
The complete grid yields
$5\!\times\!4\!\times\!8\!\times\!3\!\times\!25\!\times\!5=60{,}000$ aggregated
measurements.

\begin{figure}[t]
\centering
\includegraphics[width=\linewidth]{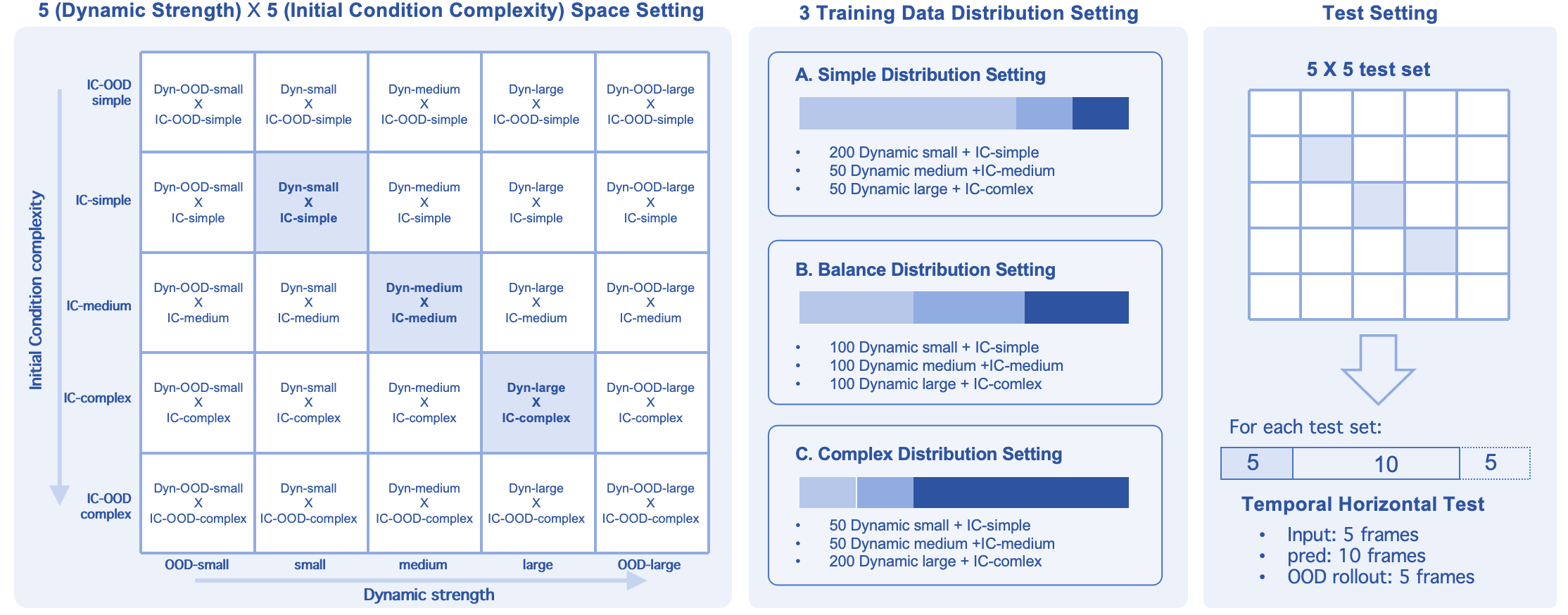}
\caption{Overview of the benchmark protocol. The evaluation space is organized as a
$5\times5$ grid over dynamic-scale setting and initial-condition complexity. The three
diagonal cells correspond to train-seen regimes, while the remaining cells test
compositional in-range shifts, dynamic-scale OOD shifts, IC-complexity OOD shifts,
and joint OOD shifts. We construct three training mixtures with different biases
toward simple, balanced, or complex regimes, while evaluating every model on the
same full $5\times5$ test grid. For each test case, models are given 5 input
frames, predict 10 in-horizon frames, and are further evaluated on 5 additional
OOD rollout frames.}
\label{fig:5x5}
\end{figure}

\begin{table}[t]
\caption{Factorized evaluation axes. The benchmark includes variation of data distribution, physics dynamics, temporal horizon, and model factors separately so that average performance can be
decomposed into interpretable capability biases.}
\vspace{5pt}
\centering
\scriptsize
\begin{tabular}{@{}p{0.16\linewidth}p{0.42\linewidth}p{0.36\linewidth}@{}}
\toprule
Axis & Values & What it tests \\
\midrule
Architecture $m$ &
DPOT, GPhyT, MORPH, MPP, Poseidon &
Architecture-dependent inductive bias \\
Variant $v$ &
Scratch and pretrained S/M/L variants &
Pretraining transfer and scaling behavior \\
Physical dynamics $p$ &
Fisher-KPP, Gray-Scott, Swift-Hohenberg, Burgers, Kolmogorov,
Kuramoto-Sivashinsky, Decay, Wave &
PDE-family and physics-regime bias \\
Training mixture $r$ &
Mix-simple, Mix-balance, Mix-complex &
Whether data coverage changes the bias pattern \\
Test cell $(d,q)$ &
$5\!\times\!5$ dynamic-scale $\times$ IC-complexity grid &
Compositional, temporal-OOD, IC-OOD, and joint-OOD generalization \\
Horizon $H$ &
1-step, 5-step, 10-step, OOD rollout, overall steps &
Short-term accuracy and longer-rollout stability \\
\bottomrule
\end{tabular}
\label{tab:eval_axes}
\end{table}

\subsection{Training mixtures}

The training distribution is treated as an experimental treatment rather than a fixed background choice. We construct three mixtures with controlled proportions
over the three train-seen diagonal regimes (\autoref{fig:5x5}). Mix-simple overweights the simple diagonal regime, Mix-balance allocates data evenly across
the three diagonal regimes, and Mix-complex overweights the complex diagonal regime. The total training-set size, model architecture, evaluation protocol, and
test cells are held fixed across mixtures. This design lets us ask whether richer or more complex training data changes the capability profile, or only changes
average error.

\subsection{Test regimes}

All trajectories are generated procedurally using APEBench and its Exponax
reference simulator~\citep{koehler2024apebench}. For each PDE and
initial-condition setting, we first simulate a dense trajectory of approximately
100 frames. The IC axis is controlled through the APEBench initial-condition
generator, whose parameters change the spatial complexity of the starting field.
The dynamic-scale axis is constructed after simulation by temporally subsampling
the dense trajectory. Consecutive frames produce the smallest apparent dynamics,
while larger frame strides produce larger state changes between prediction
steps. Thus, the dynamic labels describe the effective temporal change seen by
the forecasting model: \emph{small}, \emph{medium}, and \emph{large} are
in-range stride levels, while \emph{OOD-small} and \emph{OOD-large} are
out-of-range stride levels on the slower and faster sides.

For each PDE family, test distributions form a $5\!\times\!5$ grid over
dynamic-scale setting and IC complexity:
\[
\text{Dynamic} \!\in\! \{\text{OOD-small},\text{small},\text{medium},\text{large},\text{OOD-large}\},
\]
\[
\text{IC} \!\in\! \{\text{OOD-simple},\text{simple},\text{medium},\text{complex},\text{OOD-complex}\}.
\]
The three diagonal cells $(\text{small},\text{simple})$,
$(\text{medium},\text{medium})$, and $(\text{large},\text{complex})$ are train-seen regimes. The other cells are grouped into four interpretable shift types: \emph{Compositional ID} for unseen in-range combinations,
\emph{Dynamic OOD} for out-of-range dynamic scale, \emph{IC OOD} for
out-of-range IC complexity, and \emph{Joint OOD} when both axes are out of
range. ~\autoref{fig:5x5} shows the layout. 
Moving horizontally
changes the dynamic-scale setting while holding IC complexity fixed; moving vertically changes
IC complexity while holding the dynamic-scale setting fixed; corners combine both shifts.

\subsection{Diagnostic metrics}

Our base metric is per-sample-frame relative $L_2$ error, averaged over
test trajectories and frames:
\begin{equation}
E_H = \frac{1}{|\mathcal{S}||H|}\sum_{i\in \mathcal{S}}\sum_{t\in H}
\frac{\|\hat{u}_{i,t} - u_{i,t}\|_2}{\|u_{i,t}\|_2}.
\end{equation}
Computing the relative error per sample and frame before aggregation avoids allowing high-energy trajectories, or frames with larger absolute field
magnitude, to dominate the metric. Unless otherwise stated, non-rollout analyses use the 10-step in-horizon error. Long-horizon rollout analyses are stated
explicitly. When reporting group-level uncertainty, we use paired $95\%$ bootstrap confidence intervals obtained by resampling the $50$ test
trajectories.
Average error gives only the overall ranking; it does not show which PDE families, shifts, horizons, or model choices drive the score. We therefore define
diagnostics that normalize error along the main axes of the benchmark: PDE family, test-cell shift, rollout horizon, pretraining state, and model size
(~\autoref{tab:metrics}).

\begin{table}[t]
\caption{Diagnostic metric suite. Unless otherwise stated, all quantities use
the same prediction horizon when forming ratios. \emph{Train-seen} denotes the three
diagonal cells; \emph{Compositional ID} denotes unseen in-range combinations; and
\emph{Dynamic/IC/Joint OOD} denote out-of-range shifts along one or both axes.
For PretrainingGain, the scratch checkpoint is compared only with the
size-matched finetuned $M$ variant to avoid confounding pretraining with model
size.
$E_{\mathrm{1\text{-}step}}$ denotes the first predicted-frame error, and
$E_{\mathrm{roll}}$ denotes the explicitly stated rollout horizon error.
}
\centering
\scriptsize
\begin{tabular}{@{}p{0.16\linewidth}p{0.4\linewidth}p{0.36\linewidth}@{}}
\toprule
Metric & Definition & Interpretation \\
\midrule
Base error &
$E_H(m,v,p,r,c)$ &
Per-sample-frame relative $L_2$ error \\
PDEBias &
$E(m,v,p)/\mathrm{mean}_{p'}E(m,v,p')$ &
$>1$: model is relatively weak on PDE $p$ \\
ShiftDamage &
$E(m,v,p,c)/\mathrm{mean}_{c'\in\mathcal{C}_{\mathrm{train}}}E(m,v,p,c')$ &
$>1$: degradation relative to train-seen cells \\
RolloutAmplification &
$E_{\mathrm{roll}}/E_{\mathrm{1\text{-}step}}$ &
Long-horizon growth from short-horizon error \\
PretrainingGain &
$\big(E_{\mathrm{scr},M}-E_{\mathrm{ft},M}\big)/E_{\mathrm{scr},M}$ &
$>0$: matched-size pretraining improves accuracy \\
ModelSizeGain &
$\big(E_{\mathrm{ft},S}-E_{\mathrm{ft},s}\big)/E_{\mathrm{ft},S},\ s\in\{M,L\}$ &
$>0$: larger finetuned model improves over S \\
\bottomrule
\end{tabular}
\label{tab:metrics}
\end{table}


\section{Results}

\begin{wrapfigure}{r}{0.6\linewidth}
  \vspace{-1.2em}
  \centering
  \includegraphics[width=\linewidth]{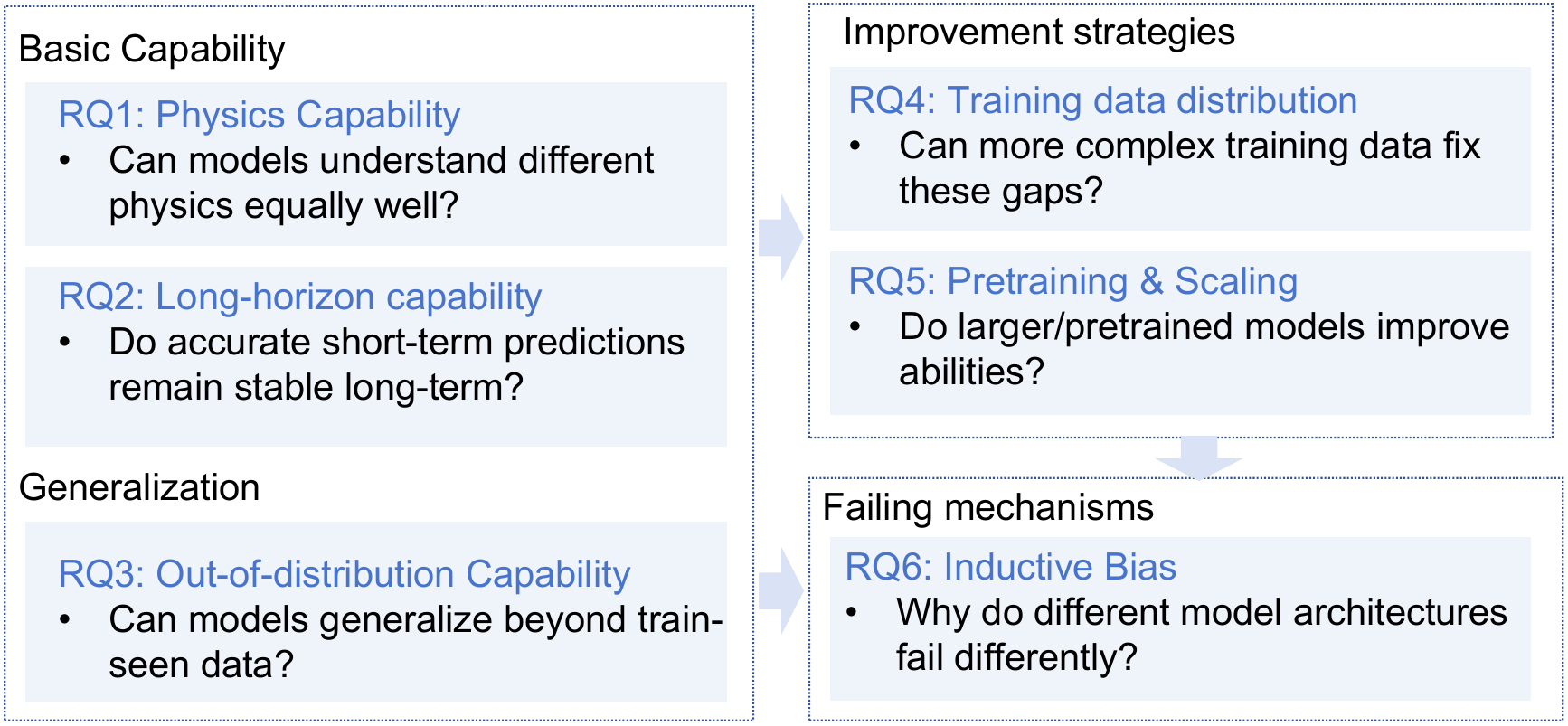}
  \vspace{-1.4em}
\end{wrapfigure}
We organize the results as a four-stage diagnosis. First, we evaluate basic capability across physical dynamics and rollout horizon (RQ1--RQ2). Second, we
test whether that capability generalizes beyond train-seen dynamic and initial-condition regimes (RQ3). Third, we ask whether common improvement
strategies---richer training mixtures, pretraining, and scaling---close the observed gaps (RQ4--RQ5). Finally, we summarize architecture-specific failure
fingerprints to explain why models with similar average errors can fail in different ways (RQ6).

\subsection{RQ1: Do physics foundation models exhibit uniform capability across physical dynamics?}

We first test whether model capability is uniform across physical dynamics. We use Mix-balance, the three train-seen diagonal cells, and the 10-step horizon to
isolate PDE-family effects from distribution-shift effects. Under this controlled setting, performance differences mainly reflect sensitivity to the physical
dynamics rather than extrapolation.

\begin{figure}[h]
    \centering
    \includegraphics[width=\linewidth]{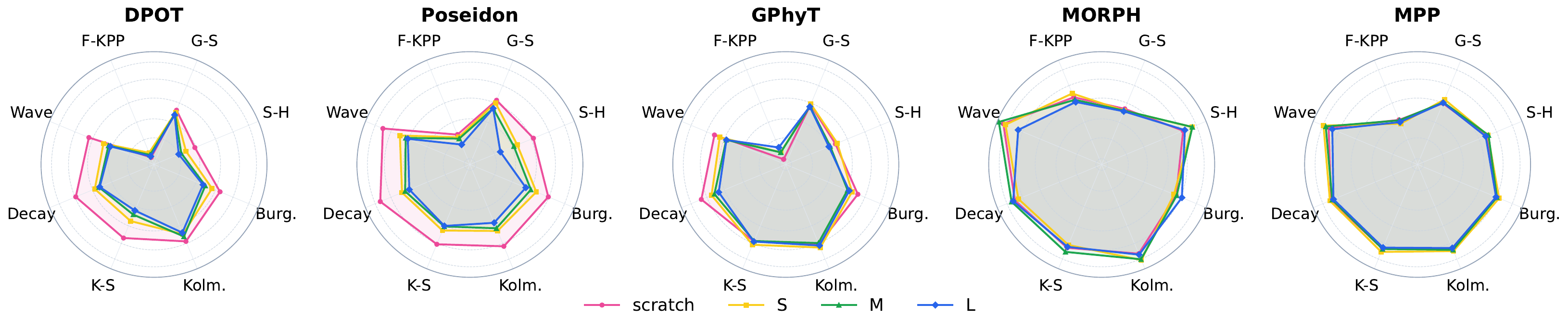}
    \caption{\textbf{Raw PDE performance.} Each plot corresponds to one model family, 
    and each curve shows one model variant: scratch, S, M, and L. 
    The eight axes denote PDE families, and the plotted value is the raw relative $L_2$ error averaged over the three train-seen cells and the 10 in-horizon predicted frames. 
    Lower values indicate better absolute accuracy. 
    The radial axis is shown on a log scale to make both low-error and high-error PDE regimes 
    visible in the same plot.}
    \label{fig:pde_raw_error_radar}
\end{figure}

\begin{figure}[h]
    \centering
    \includegraphics[width=\linewidth]{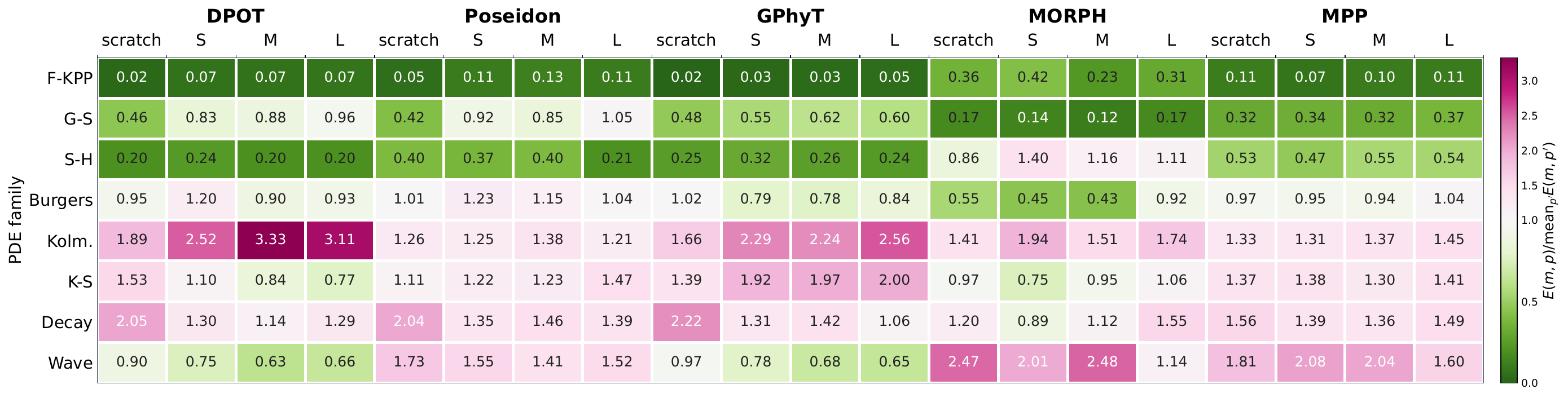}
    \caption{\textbf{PDEBias.}
    Rows correspond to PDE families and columns correspond to model variants, grouped by architecture and variant type.
    Each cell reports the within-model normalized error
    $E(m,p)/\mathrm{mean}_{p'}E(m,p')$, computed using relative $L_2$ error averaged over the three train-seen cells and the 10 in-horizon predicted frames.
    Values below $1$ indicate PDE families on which a model is relatively stronger than its own average, 
    while values above $1$ indicate relatively harder PDE families.
    Unlike the raw radar plot, this normalization removes each model's overall error scale and highlights 
    whether its capability is balanced across PDE families.}
    \label{fig:pde_bias_strip}
\end{figure}

\autoref{fig:pde_raw_error_radar} shows that raw accuracy is highly PDE-dependent even under train-seen conditions. Fisher--KPP is the easiest
regime for most variants: its median relative $L_2$ error across the 20 variants is $0.011$, and it is the lowest-error PDE for DPOT, GPhyT, MPP, and Poseidon.
Gray--Scott is also consistently low-error and is the easiest PDE for all MORPH variants. In contrast, the largest errors concentrate on higher-variation
dynamics. Kolmogorov is the hardest PDE for most DPOT and GPhyT variants, while Wave dominates the failures of MORPH and MPP; the worst case is MORPH-M on Wave
($E=1.33$). Thus, even on train-seen distributions, model performance can vary by one to two orders of magnitude across PDE families.

Raw error shows absolute performance, but it is not sufficient to characterize whether a model has a balanced capability profile across PDE families. 
A model with low overall error can still be disproportionately strong on some physical
regimes and weak on others, while a less accurate model may perform more uniformly.
Therefore, to evaluate physical-regime bias within each model, we compute a within-model normalized PDEBias:

\[
    \mathrm{PDEBias}(m,p)
    =
    \frac{E(m,p)}{\mathrm{mean}_{p'}E(m,p')},
\]
where $E(m,p)$ is the relative $L_2$ error of model variant $m$ on PDE family $p$, which compares
each PDE against the model's own average PDE error, separating a model's PDE-specific preference from its overall accuracy level. Values below $1$
indicate that the PDE is relatively easier for that model, while values above $1$ indicate that it is relatively harder.
As shown in ~\autoref{fig:pde_bias_strip}, 
the PDE dependence remains.
Most variants are relatively strong on Fisher--KPP, Gray--Scott, and Swift--Hohenberg, while Kolmogorov,
Kuramoto--Sivashinsky, Decay, and Wave frequently appear as above-average difficulty regimes.
We can also see the bias pattern is model architecture dependent. MORPH shows a stronger relative
weakness on Wave than DPOT and GPhyT, whereas DPOT and GPhyT show stronger relative difficulty on Kolmogorov dynamics. Scratch, S, M, and L variants within
the same architecture often preserve similar column-wise trends, suggesting that pretraining and scaling modify the magnitude of PDE-specific errors but do not
fully remove the underlying physical-regime preference.
Together, the raw-error and normalized PDEBias results show that model capability varies strongly across physical dynamics and current physics foundation models exhibit systematic PDE-family-dependent capability bias even under train-seen conditions.

\begin{findingbox}
\textbf{Finding 1.}
Physics foundation models exhibit systematic physical-regime bias. Their capability is relatively stronger on several
reaction--diffusion and pattern-forming regimes, but weaker on high-variation or wave-like regimes such as Kolmogorov, Kuramoto--Sivashinsky, Decay, and Wave.
Pretraining and scaling change the magnitude of errors, but do not fully remove these PDE-specific capability profiles.
\end{findingbox}

\subsection{RQ2: Do physics foundation models maintain capability across prediction horizons?}\label{sec:rq2}


We next examine horizon-dependent forecasting capability. Under the same controlled setting as RQ1, we compare raw frame-wise error with normalized error
amplification across prediction horizons, which separates immediate accuracy from temporal stability, for example, models with similar short-horizon errors may accumulate
errors at very different rates, and models with weaker early predictions may be
more stable over longer rollouts. Let
\[
    \mathrm{RolloutAmplification}_{10}(m,p)
    =
    \frac{E_{\mathrm{10\text{-}step}}(m,p)}
    {E_{\mathrm{1\text{-}step}}(m,p)} ,
\]
where $E_{\mathrm{1\text{-}step}}$ is the first predicted-frame error and
$E_{\mathrm{10\text{-}step}}$ is the average error over the 10 in-horizon predicted frames. This ratio measures how strongly a model's own initial error is amplified over the standard prediction horizon.

\begin{figure}[h]
    \centering
    \includegraphics[width=\linewidth]{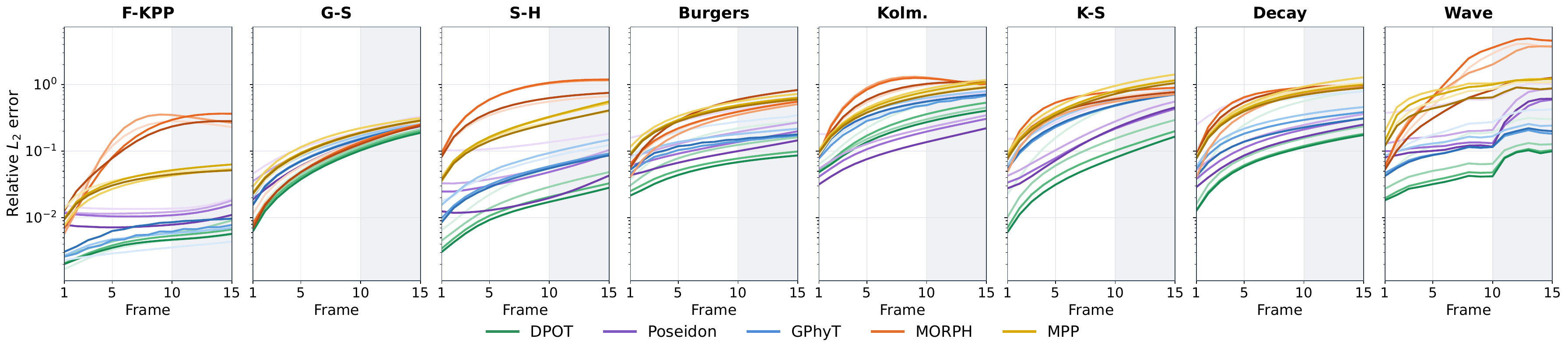}
    \caption{\textbf{PDE-wise rollout error growth.}
    Each subplot is one PDE family. Curves show the frame-wise relative $L_2$ error of all model variants, with colors grouped by architecture and shade
indicating scratch/S/M/L variants. The gray region marks frames beyond the 10-step training horizon. Errors generally grow with prediction time, but
    the growth pattern depends strongly on both PDE family and model variant.}
    \label{fig:rq4_rollout_curves}
\end{figure}

\begin{figure}[h]
    \centering
    \includegraphics[width=\linewidth]{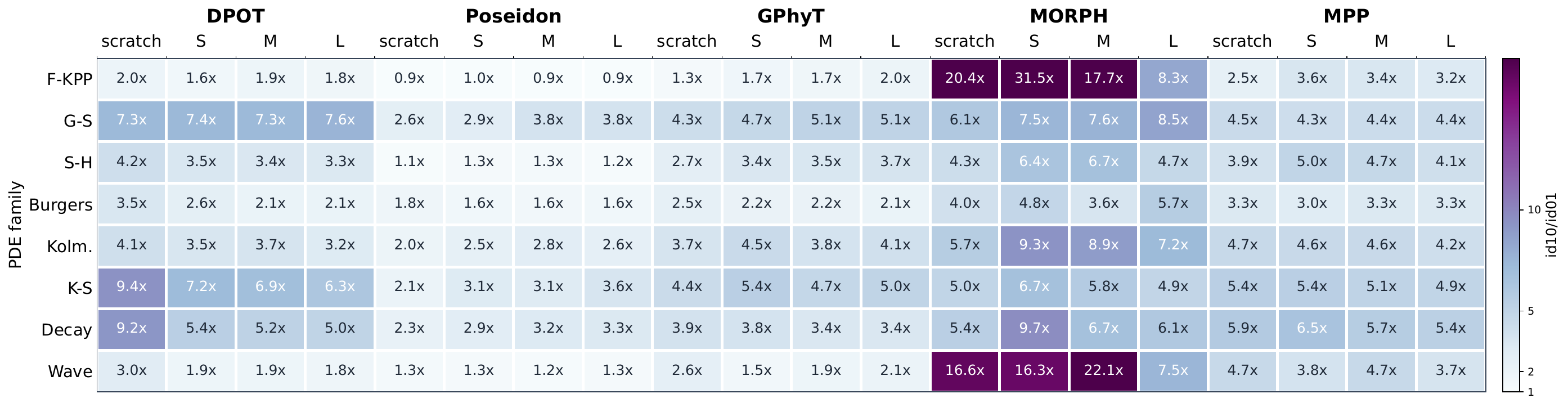}
    \caption{\textbf{10-step in-horizon error amplification by model variant and PDE family.}
    Rows correspond to PDE families and columns correspond to model variants,
    grouped by architecture and variant type. Each cell reports
    $E_{\mathrm{10\text{-}step}}/E_{\mathrm{1\text{-}step}}$, where
    $E_{\mathrm{1\text{-}step}}$ is the first predicted-frame error and
    $E_{\mathrm{10\text{-}step}}$ is the average error over the 10 in-horizon
    predicted frames. Larger values indicate stronger error accumulation within
    the standard prediction horizon.}
    \label{fig:rq4_amplification_strip}
\end{figure}

~\autoref{fig:rq4_rollout_curves} shows systematic but uneven temporal degradation. Errors usually increase with prediction time, especially after the
gray OOD rollout region begins, but the growth curves differ substantially across architectures and PDE families. Some variants grow smoothly, whereas
others show sharp late-stage amplification. This indicates that horizon-dependent capability has two components: the absolute error reached at later frames and the rate at which early errors are amplified.

~\autoref{fig:rq4_amplification_strip} quantifies how much each model's first-frame error is amplified within the standard prediction horizon.
Error amplification is strongly architecture- and PDE-dependent. Poseidon has the lowest mean amplification ($2.09\times$) despite
having the largest mean first-frame error among the five architectures ($0.072$). In contrast, DPOT has the lowest mean first-frame error ($0.019$) but
a larger amplification factor ($4.34\times$). MORPH illustrates the separation between short- and long-horizon behavior: its first-frame error is
competitive on several PDEs, including Gray--Scott, Burgers, and Wave, where it outperforms Poseidon and MPP on immediate prediction. However, these early
advantages often do not persist. MORPH has the largest mean amplification ($9.11\times$), with extreme cells such as MORPH-S on Fisher--KPP
($31.5\times$) and MORPH-M on Wave ($22.1\times$). Across all 160 model--PDE test, first-frame error is strongly
correlated with raw 10-step error (Spearman $\rho=0.86$), but almost uncorrelated with 10-step amplification (Spearman $\rho=0.04$). Therefore,
early-frame accuracy and temporal stability capture different aspects of forecasting capability.

\begin{findingbox}
\textbf{Finding 2.}
Physics foundation models exhibit horizon-dependent capability bias. Competitive early-frame accuracy does not necessarily translate into stable longer-horizon behavior: some models amplify errors rapidly, while others maintain lower growth despite weaker short-term accuracy. Current models therefore learn
temporal dynamics unevenly across horizons rather than stable forecasting behavior that transfers across time.
\end{findingbox}

\subsection{RQ3: Do physics foundation models generalize beyond train-seen distributions?}\label{sec:rq3_shift}

After analyzing physical-regime bias and horizon-dependent behavior, we ask how models generalize when the test condition moves away from the train-seen cells.
Here, distribution shift is not a binary ID/OOD label. The $5\times5$ grid separates two physically meaningful directions of shift: dynamic scale and
initial-condition complexity. This lets us ask which direction creates the larger generalization gap, and whether the answer depends on the PDE family.
We measure this gap with ShiftDamage, which normalizes each shifted cell by the same model's train-seen performance on the same PDE. For each model variant and
PDE family, we define
\[
    \mathrm{ShiftDamage}(m,p,c)
    =
    \frac{E(m,p,c)}
    {\mathrm{mean}_{c'\in\mathcal{C}_{\mathrm{train}}}E(m,p,c')},
\]
where $c$ is a test cell and $\mathcal{C}_{\mathrm{train}}$ contains the three
train-seen diagonal cells. ShiftDamage measures relative degradation from a model's own train-seen baseline. A value
near $1$ means that the shifted cell is no harder than the train-seen reference for that model--PDE pair, while values above $1$ indicate a robustness gap under
the corresponding physical shift.

\begin{figure}[h]
    \centering
    \includegraphics[width=\linewidth]{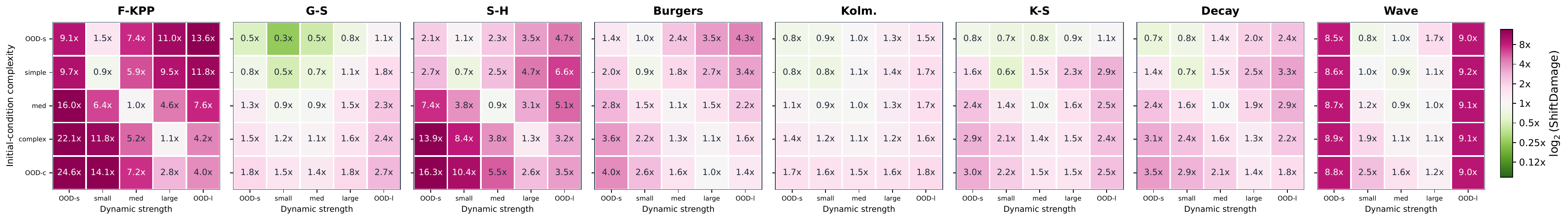}
    \caption{\textbf{PDE-wise $5\!\times\!5$ ShiftDamage grids under Mix-balance.}
    Each panel corresponds to one PDE family, computed at the 10-step horizon.
    Columns increase dynamic strength from OOD-small to OOD-large, and rows increase initial-condition complexity
    from OOD-simple to OOD-complex. Each cell reports ShiftDamage averaged over all evaluated model variants, with each variant normalized by its own
    train-seen diagonal baseline. Colors use
    $\log_2(\mathrm{ShiftDamage})$: $1\times$ maps to zero/white, values below
    $1\times$ are green, and values above $1\times$ are red.}
    \label{fig:rq3_pde_5x5}
\end{figure}

\autoref{fig:rq3_pde_5x5} shows that distribution shift is organized by the direction of the physical change. Dynamic-scale shifts produce the clearest
relative error growth. For Wave, the two Dynamic-OOD columns remain around $9\times$ across most IC settings, while the in-range dynamic cells stay
much closer to the train-seen baseline. Fisher--KPP and Swift--Hohenberg show a similar sensitivity to dynamic-scale extrapolation. In contrast, Gray--Scott,
Kolmogorov, and Kuramoto--Sivashinsky contain more near-baseline or reduced-error cells, especially around simpler IC settings. Thus, moving away
from the train-seen diagonal does not have a single effect: depending on the PDE and shift direction, error can increase sharply, remain stable, or even decrease.

Appendix ~\ref{fig:app_rq3_dpot_variant_5x5}--\ref{fig:app_rq3_mpp_variant_5x5} provide the model-resolved view of this pattern. The broad PDE-dependent layout
remains visible across architectures, but the contrast between train-seen and shifted cells changes with pretraining state and size. Larger or pretrained
variants often reduce error on train-seen cells more than on shifted cells, which makes the normalized robustness gap more visible. Across Dynamic-OOD and
Joint-OOD groups, mean ShiftDamage increases from $4.22\times$ for S variants to $6.28\times$ for L variants, while scratch variants average $3.24\times$.
Because this metric is normalized by each variant's own train-seen baseline, the increase indicates uneven transfer of train-seen gains to shifted regimes. The
group-level summary in
\autoref{fig:app_rq3_shift_groups} gives the same message: median ShiftDamage is higher for Dynamic-OOD ($1.9\times$) and Joint-OOD
($1.6\times$) than for IC-OOD ($1.3\times$), and their much larger means
($4.8\times$ for both groups).

\begin{findingbox}
\textbf{Finding 3.}
Physics foundation models do not generalize consistently beyond train-seen
distributions. Their robustness boundary is structured by both PDE family and
shift direction. Dynamic-scale and joint shifts cause the largest relative error
growth, especially for Fisher--KPP, Swift--Hohenberg, and Wave, whereas many
IC-only shifts remain close to the train-seen baseline or even reduce relative
error.
\end{findingbox}

\subsection{RQ4: How does the training distribution reshape model capability?}\label{sec:rq4_training}

The previous sections show that model capability depends on physical regime, prediction horizon, and test shift. We next ask how much of this conditional
behavior can be changed by the training distribution. The three mixtures have the same total scale but allocate data differently across simple, balanced, and
complex regimes. This separates two effects: whether a more complex mixture lowers absolute error, and whether those gains carry over to shifted
test cells.
We use the same normalized ShiftDamage as in RQ3, but compute the train-seen
baseline separately for each training mixture:
\[
\mathrm{ShiftDamage}(t,m,p,c)
=
\frac{E(t,m,p,c)}
{\mathrm{mean}_{c'\in \mathcal{C}_{\mathrm{train}}}E(t,m,p,c')},
\]
where $t$ is the training mixture, $m$ is the model variant, $p$ is the PDE family, and $c$ is a test cell. This normalization asks whether a mixture
improves unseen cells relative to the competence it achieves on its own train-seen cells.

\begin{figure}[h]
\centering
\includegraphics[width=0.7\linewidth]{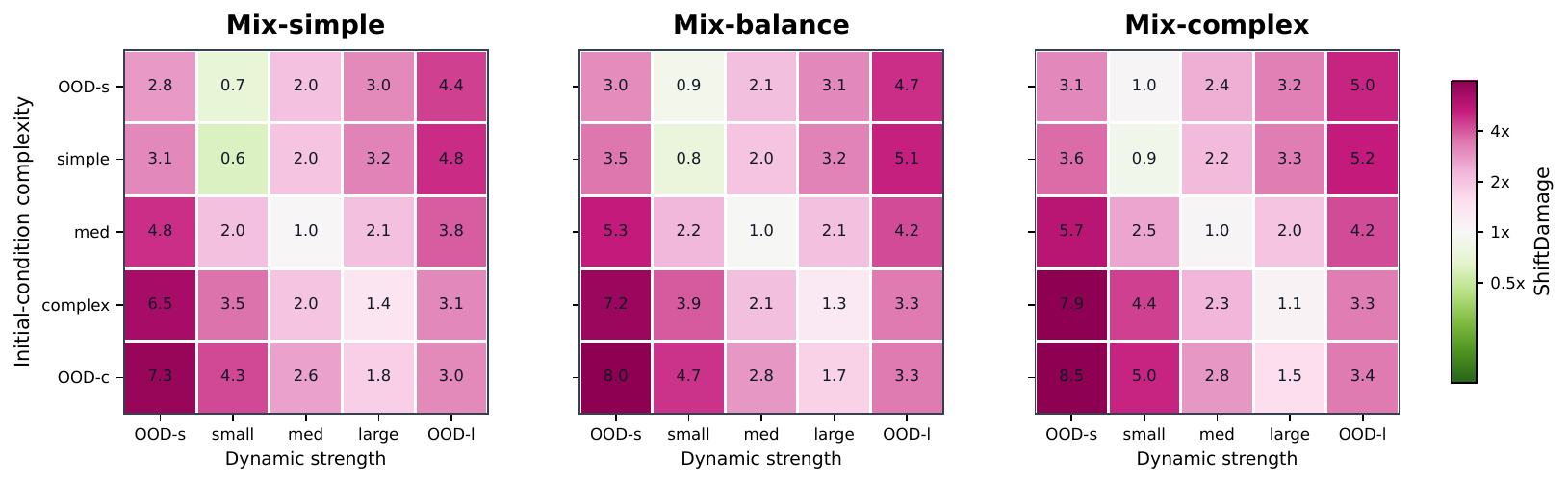}
\caption{\textbf{Training-mixture effect on the $5\times5$ ShiftDamage grid.}
Each panel corresponds to one training mixture. Rows vary initial-condition
complexity and columns vary dynamic scale. Each cell reports mean normalized
ShiftDamage averaged over model variants and PDE families, using the
10-step relative $L_2$ error. Colors use
$\log_2(\mathrm{ShiftDamage})$: $1\times$ maps to white, values below
$1\times$ are green, and values above $1\times$ are red.}
\label{fig:rq4_mix_5x5}
\end{figure}

\begin{table}[h]
\caption{\textbf{Raw accuracy and normalized robustness by training mixture.}
Values are averaged over model variants, PDE families, and cells in each shift
group. Raw error measures absolute relative $L_2$ error, while ShiftDamage
normalizes each mixture by its own train-seen diagonal baseline.}
\centering
\small
\resizebox{\linewidth}{!}{%
\begin{tabular}{lcccccc}
\toprule
\multirow{2}{*}{Shift group}
& \multicolumn{3}{c}{Raw error}
& \multicolumn{3}{c}{ShiftDamage} \\
\cmidrule(lr){2-4}\cmidrule(lr){5-7}
& Mix-simple & Mix-balance & Mix-complex
& Mix-simple & Mix-balance & Mix-complex \\
\midrule
Train-seen & 0.240 & 0.225 & 0.224 & 1.00 & 1.00 & 1.00 \\
Comp. ID & 0.298 & 0.282 & 0.283 & 2.46 & 2.57 & 2.77 \\
Dyn-OOD & 0.436 & 0.428 & 0.424 & 4.33 & 4.75 & 4.98 \\
IC-OOD & 0.304 & 0.287 & 0.288 & 2.40 & 2.55 & 2.66 \\
Joint-OOD & 0.429 & 0.419 & 0.416 & 4.36 & 4.75 & 4.96 \\
\bottomrule
\end{tabular}
}

\label{tab:rq4_training_mixture}
\end{table}

\autoref{tab:rq4_training_mixture} separates absolute accuracy from relative robustness. In raw error, Mix-complex improves over Mix-simple in every shift
group, including train-seen cells ($0.240\!\rightarrow\!0.224$), IC-OOD cells ($0.304\!\rightarrow\!0.288$), and Joint-OOD cells
($0.429\!\rightarrow\!0.416$). Thus, more complex training data does increase average predictive capability. The improvement, however, is spatially selective.
\autoref{fig:rq4_mix_5x5} shows that Mix-complex improves high-dynamic cells most clearly relative to Mix-balance, including Dynamic-large $\times$
IC-complex ($9.5\%$ lower error), Dynamic-large $\times$ IC-OOD-complex ($8.7\%$ lower), and Dynamic-OOD-large $\times$ IC-complex ($5.8\%$ lower). The
same mixture increases error in several small-dynamic cells, including Dynamic-small $\times$ IC-simple ($7.6\%$ higher error) and Dynamic-small
$\times$ IC-OOD-simple ($8.0\%$ higher). Training data therefore redistributes capability across the test grid rather than improving every region together.

The ShiftDamage columns show that these raw gains do not close the relative robustness gap. Mix-complex has larger normalized damage than Mix-simple in
every non-train group: $2.46\!\rightarrow\!2.77$ on compositional cells, $4.33\!\rightarrow\!4.98$ on Dynamic-OOD, $2.40\!\rightarrow\!2.66$ on IC-OOD,
and $4.36\!\rightarrow\!4.96$ on Joint-OOD. Since the baseline is recomputed within each mixture, this means that train-seen gains transfer unevenly to
shifted regimes. Dynamic-OOD and Joint-OOD remain the largest gaps under all three mixtures, so changing the training distribution reshapes model strengths
but does not remove the conditional generalization behavior observed in RQ1--RQ3.

\begin{findingbox}
\textbf{Finding 4.}
More complex training data shifts model strengths across regimes, but does not consistently improve generalization. 
It improves absolute accuracy in selected regions, especially large-dynamic regimes, but the gains are uneven and do not
carry over across physical shifts. The persistent Dynamic-OOD and Joint-OOD gaps show that more complex training data alone is insufficient to
produce robust OOD capability across the grid.

\end{findingbox}

\subsection{RQ5: How do pretraining and scaling reshape model capability?}\label{sec:rq5_model_interventions}

After studying data-side changes in RQ4, we now examine two model-side interventions: pretraining and scaling. The goal is not only to test whether
they reduce error on average, but to ask whether the resulting gains are stable across physical regimes and test conditions. We therefore use controlled
comparisons: pretraining is evaluated at matched size by comparing scratch-$M$ and pretrained-$M$, while scaling is evaluated within pretrained variants using
$S$ as the baseline. Here, we define use PretrainingGain and ModelSizeGain:

\[
\mathrm{PretrainingGain}(a,p)
=
\frac{E_{\mathrm{scratch},M}(a,p)-E_{\mathrm{ft},M}(a,p)}
{E_{\mathrm{scratch},M}(a,p)}\times 100.
\]
$E(a,p)$ is the relative $L_2$ error for architecture
$a$ and PDE family $p$, averaged over the full grid. Positive values indicate beneficial transfer, while negative values indicate negative transfer. Scaling
is evaluated only among pretrained variants by using the $S$ model as the within-architecture baseline:
\[
\mathrm{ModelSizeGain}(a,p,s)
=
\frac{E_{\mathrm{ft}}(a,p,S)-E_{\mathrm{ft}}(a,p,s)}
{E_{\mathrm{ft}}(a,p,S)}\times 100,
\qquad s\in\{S,M,L\}.
\]
By definition, the $S$ column is zero; positive $M/L$ values mean that a larger
model helps, while negative values mean inverse scaling.

\begin{figure}[t]
\centering
\includegraphics[width=0.5\linewidth]{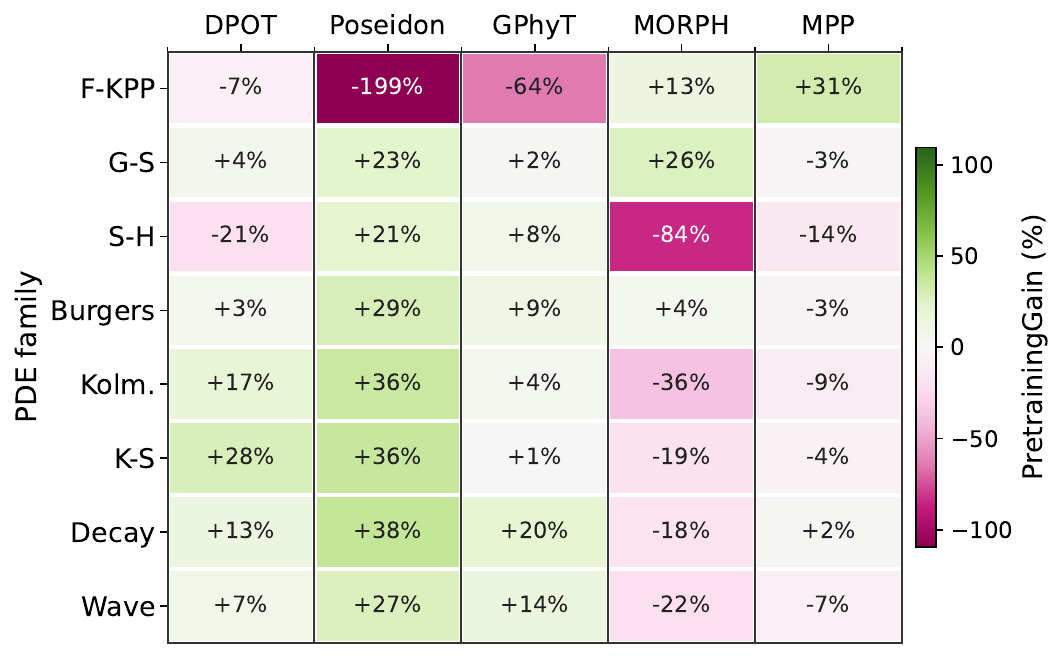}
\caption{\textbf{Matched-size PretrainingGain under Mix-balance.}
Rows correspond to PDE families and columns correspond to architectures. Each cell reports the percentage error reduction of pretrained-$M$ relative to
scratch-$M$ for the same architecture, averaged over the full $25$-cell grid at the 10-step horizon. Green indicates beneficial transfer and pink
indicates negative transfer.}
\label{fig:rq5_pretraining_gain}
\end{figure}

\autoref{fig:rq5_pretraining_gain} shows that pretraining reshapes capability in a regime-dependent way. It improves some architecture--PDE pairs, but does not
produce a consistently better model. Across the $5\times8$ architecture--PDE cells, $37.5\%$ show negative transfer and the mean gain is slightly negative
($-2.3\%$). The effect is strongly conditional: Poseidon-M improves Decay by $+38\%$ but degrades Fisher--KPP by $-199\%$, while GPhyT-M degrades
Fisher--KPP by $-64\%$ but improves Decay by $+20\%$. Thus, pretraining appears to transfer regime-specific preferences rather than a universally useful physics
prior.

\begin{figure}[t]
\centering
\includegraphics[width=\linewidth]{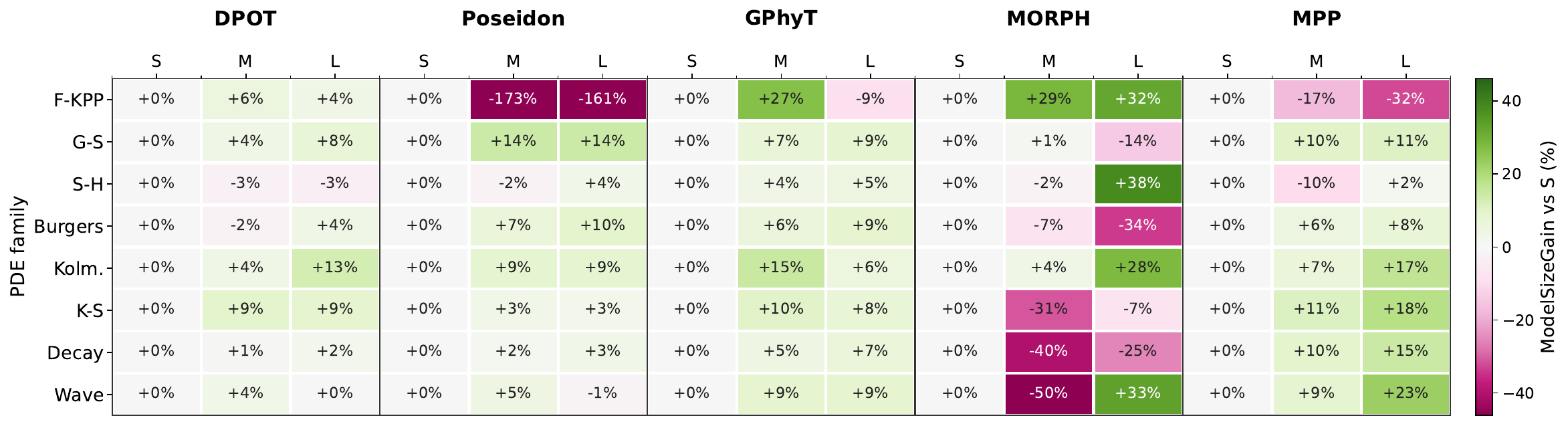}
\caption{\textbf{ModelSizeGain under Mix-balance.}
Rows correspond to PDE families and columns correspond to model variants, grouped by architecture and finetuned size. Each cell reports the percentage
error reduction relative to the $S$ finetuned model of the same architecture and PDE. The $S$ column is therefore zero. Green indicates that the larger model
improves over $S$; pink indicates inverse scaling.}
\label{fig:rq5_model_size_gain}
\end{figure}

\autoref{fig:rq5_model_size_gain} shows that scaling has the same selective character. Larger models often help: $L$ is the best size in 23 of the 40
architecture--PDE pairs. However, scaling is not monotonic. Only $45.0\%$ of architecture--PDE pairs improve monotonically from $S$ to $M$ to $L$, and
$25.0\%$ of $M/L$ cells are worse than the $S$ baseline. For example, MORPH-L improves Wave by $+34\%$ and Swift--Hohenberg by $+38\%$, but MORPH-M is worse
than S on Wave ($-50\%$), and Poseidon-M/L are much worse than S on Fisher--KPP ($-173\%$ and $-161\%$). Scaling therefore reallocates capability across regimes
rather than improving all regimes together.

\begin{findingbox}
\textbf{Finding 5.}
Pretraining and scaling produce selective capability gains. Pretraining is not universally beneficial: even under a matched-size comparison, $37.5\%$ of
architecture--PDE pairs show negative transfer. Scaling is also non-monotonic. Although the large variant is best in 23 of 40 pairs, $25.0\%$ of larger-model
cells are worse than the small baseline.
\end{findingbox}

\subsection{RQ6: Do physics foundation models fail through different mechanisms?}\label{sec:rq6_mechanisms}

The previous RQs show that failures arise along several axes: physical-regime bias, distribution shift, rollout stability, pretraining transfer, and scaling.
We now ask whether architectures share the same weakness profile, or whether each architecture fails through a different combination of mechanisms. To make
these profiles comparable, we aggregate five diagnostics under Mix-balance: \emph{PDE bias severity} as the mean absolute log PDEBias across PDE families,
\emph{OOD ShiftDamage} as the mean normalized degradation over non-train-seen cells, \emph{RolloutAmplification} as
$E_{\mathrm{rollout\_ood5}}/E_{\mathrm{id01}}$ on the three train-seen cells, \emph{negative transfer rate} from matched scratch-$M$ versus pretrained-$M$,
and \emph{inverse scaling rate} from pretrained $M/L$ versus pretrained $S$.

\begin{figure}[t]
\centering
\includegraphics[width=0.6\linewidth]{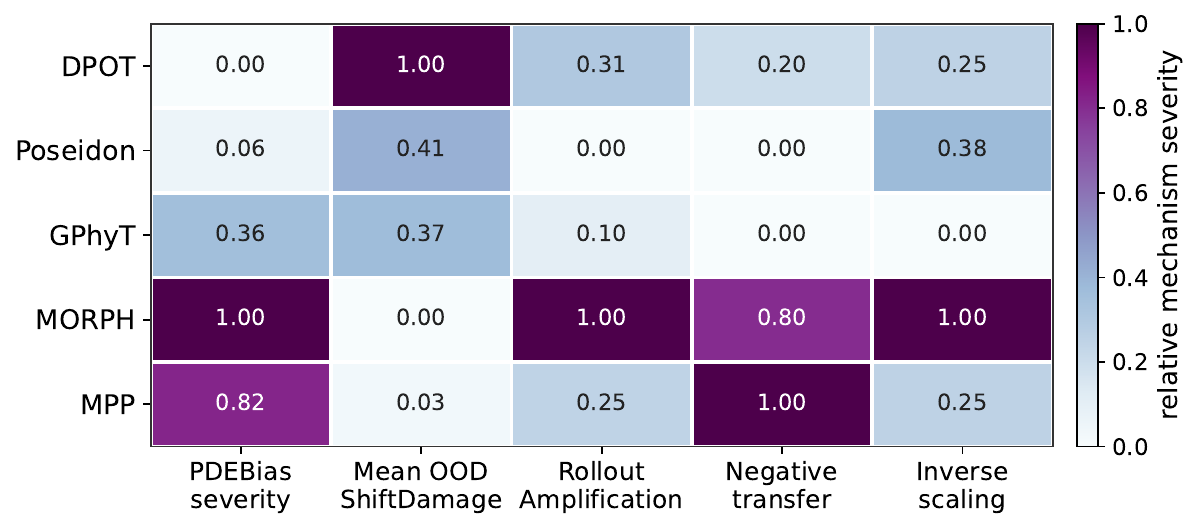}
\caption{\textbf{Architecture-level failure-mechanism fingerprint.}
Rows correspond to architectures and columns correspond to failure dimensions derived from the diagnostics above. Values are min--max normalized within each
column, so darker cells indicate that an architecture is relatively more affected by that failure mode.}
\label{fig:rq6_failure_fingerprint}
\end{figure}

\autoref{fig:rq6_failure_fingerprint} shows that architectures fail in qualitatively different ways. DPOT is most sensitive to distribution shift: it
has the largest normalized OOD degradation ($8.66\times$), even though its PDE bias severity is the lowest among the five models. MORPH is dominated by
temporal and scaling instability, with the largest rollout amplification ($22.8\times$) and the highest inverse-scaling rate ($56.2\%$). MPP shows a
different profile: it has the highest matched-size negative-transfer rate ($75.0\%$), but its temporal amplification is much smaller than MORPH's.
Poseidon and GPhyT are more balanced in this fingerprint, but still have clear weaknesses; Poseidon shows scaling sensitivity, while GPhyT remains
affected by PDE and rollout variation. These differences indicate that architecture matters not only through average accuracy, but through the type of
failure it induces.

\begin{findingbox}
\textbf{Finding 6.}
Architectures fail through different combinations of mechanisms rather than along a single accuracy axis. DPOT is most sensitive to distribution shift,
MORPH to rollout and scaling instability, and MPP to pretraining negative
transfer, while Poseidon and GPhyT show more balanced but still uneven
profiles. Thus, improving physics foundation models requires addressing
architecture-specific failure modes, not only lowering average error.
\end{findingbox}

\section{Discussion}\label{sec:discussion}

\paragraph{What current models are good for.}
Our results show that the learned competence of physical foundation model is conditional, which is different from the usual foundation-model promise of broad
reuse across tasks and regimes~\citep{bommasani2021opportunities}. In our benchmark, the same model can be strong on one PDE family, one temporal horizon,
or one shift direction, and weak on another. For practical use, this means that model choice should depend on the target regime rather than on a single averaged
score. A model that is attractive for smooth dissipative dynamics may be a poor choice for wave-like or high-variation dynamics; a model with low short-horizon
error may still be unsuitable for long rollouts. The useful object is therefore not just a model ranking, but a capability profile: where the model is accurate,
where it degrades, and which failure mode is most relevant to the intended forecasting problem.

\paragraph{What data, pretraining, and scale buy.}
Much of foundation-model progress has been driven by larger models, larger datasets, and pretraining at scale~\citep{kaplan2020scaling,hoffmann2022training}.
Our results show that these levers also matter for physics forecasting, but their effects are selective. More complex training mixtures improve raw accuracy in
targeted regions, especially cells with stronger dynamics, yet they leave large dynamic-scale and joint-shift gaps. Pretraining is also selective: it improves
some architecture--PDE pairs, but can transfer regime preferences that hurt other pairs. Scaling behaves similarly. Larger variants are often better, but the gains
are not monotonic and can reverse on particular PDE families or test cells. These patterns are consistent with a broader lesson from shortcut and simplicity-bias
studies: high average accuracy does not guarantee that the model has learned the intended transferable structure~\citep{geirhos2020shortcut,shah2020pitfalls}.

The main bottleneck exposed by the benchmark is not only insufficient data or insufficient parameter count. The hard cases are organized: dynamic-scale shifts
remain difficult, long-horizon errors grow differently across architectures, and pretraining or scaling can sharpen an existing regime preference instead of
removing it. This suggests that the next improvement step should focus on how physical knowledge is represented and transferred. The
architecture-level fingerprints also argue against a single universal fix: a model limited by rollout instability needs a different intervention from one
limited by OOD sensitivity, negative transfer, or inverse scaling.

\paragraph{Limitations and next steps.}
This paper focuses on model behavior to show where current models succeed and fail, and how those failures change with the physical regime, data
mixture, pretraining state, and model size. It does not yet explain what the models have learned internally. We see this as the next question. The error
patterns should be tied back to physical quantities. More importantly, future analysis should ask whether these models can infer the
governing regime from context. Given a few input frames, can the model recognize
the relevant physical knowledge, and then use that knowledge outside the regimes it was trained on?
Answering this would move beyond next-frame accuracy and toward
whether physics foundation models learn reusable physical concepts.

\section{Conclusion}\label{sec:concl}
This paper studies whether current physics foundation models learn broadly generalizable physical capability. Across 8 physical dynamics, 3 training mixtures, 25 test regimes, and 5 model architectures, we find that their capability is highly conditional. 
Models show systematic biases across PDE families, temporal horizons, distribution shifts, training mixtures, pretraining states, model
sizes, and architectures. More complex data, pretraining, and scaling can improve selected regimes, but they do not reliably remove these biases or produce consistent
OOD robustness. These results suggest that the next step for physics foundation models is not only to increase data scale or model size, but to learn physical
representations that transfer more reliably across regimes, temporal scales, and distribution shifts.

\bibliography{iclr2026_conference}
\bibliographystyle{iclr2026_conference}

\appendix

\section{Appendix: Benchmark Design and Dataset Generation}\label{app:data}

All benchmark trajectories are generated on two-dimensional periodic domains
using the APEBench/Exponax pseudospectral solvers~\citep{koehler2024apebench}.
Unless otherwise stated, each stored group has shape
\((N_{traj},100,C,64,64)\), where \(C\) is the number of physical channels.
For every PDE we keep five initial-condition families ordered by complexity:
\texttt{IC-OOD-simple}, \texttt{IC-simple}, \texttt{IC-medium},
\texttt{IC-complex}, and \texttt{IC-OOD-complex}. The training split contains
200 trajectories for each in-distribution family \texttt{IC-simple},
\texttt{IC-medium}, and \texttt{IC-complex}; validation and test contain 50
trajectories for each of the five families.

The dynamic-scale axis is constructed from these 100-frame trajectories by
temporal subsampling. Each evaluation sequence contains 20 frames. The five
dynamic settings use raw-frame strides \(1,2,3,4,5\), corresponding to
\texttt{Dynamic-OOD-small}, \texttt{Dynamic-small}, \texttt{Dynamic-medium},
\texttt{Dynamic-large}, and \texttt{Dynamic-OOD-large}, respectively. For
example, the stride-1 setting uses raw frames \(0,1,\ldots,19\), while the
stride-5 setting uses \(0,5,\ldots,95\). This changes the effective temporal
step seen by the forecasting model while keeping the PDE, spatial resolution,
and IC family fixed. Figure panels below use one test trajectory per family,
with rows ordered by IC complexity and columns showing ten uniformly sampled
times.

\subsection{Gray--Scott}\label{app:data-gray-scott}

\begin{figure}[h]
    \centering
    \includegraphics[width=\linewidth]{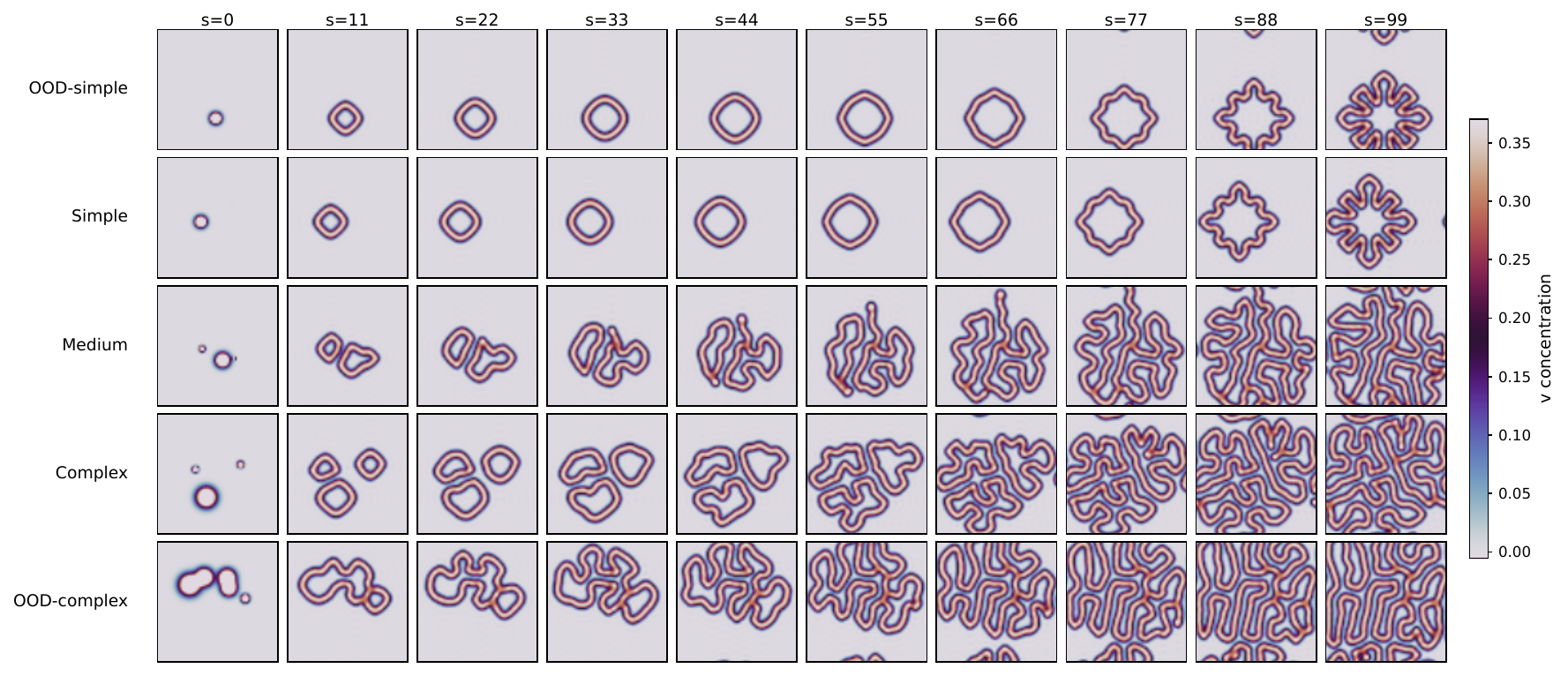}
    \caption{Gray--Scott examples. Rows are IC families ordered from OOD-simple to OOD-complex; columns are uniformly sampled times.}
    \label{fig:pde-gray-scott}
\end{figure}

The Gray--Scott system models two reacting and diffusing chemical species \(u\) and \(v\):
\begin{align}
    \partial_t u &= D_u \Delta u + F(1-u) - u v^2, \\
    \partial_t v &= D_v \Delta v - (F+k)v + u v^2 .
\end{align}
We use the Exponax Gray--Scott stepper with periodic boundary conditions, domain extent \(L=1.35\), resolution \(64\times 64\), solver step \(\Delta t=3.5\), and 14 internal solver steps per stored frame. Thus adjacent saved frames are separated by \(49.0\) time units. The parameters are \(D_u=2\times10^{-5}\), \(D_v=1\times10^{-5}\), feed rate \(F=0.044\), and kill rate \(k=0.06\). The two stored channels are the \(u\) and \(v\) concentrations; \autoref{fig:pde-gray-scott} visualizes the \(v\) channel. Initial conditions are Gaussian-blob perturbations with \(\sigma\in[0.002,0.05]\), center fraction 0.5, and no additive noise. Complexity is controlled by the number of blobs: 1, 2, 3, 4, and 5 for OOD-simple, simple, medium, complex, and OOD-complex respectively.

\subsection{Wave}\label{app:data-wave}

\begin{figure}[h]
    \centering
    \includegraphics[width=\linewidth]{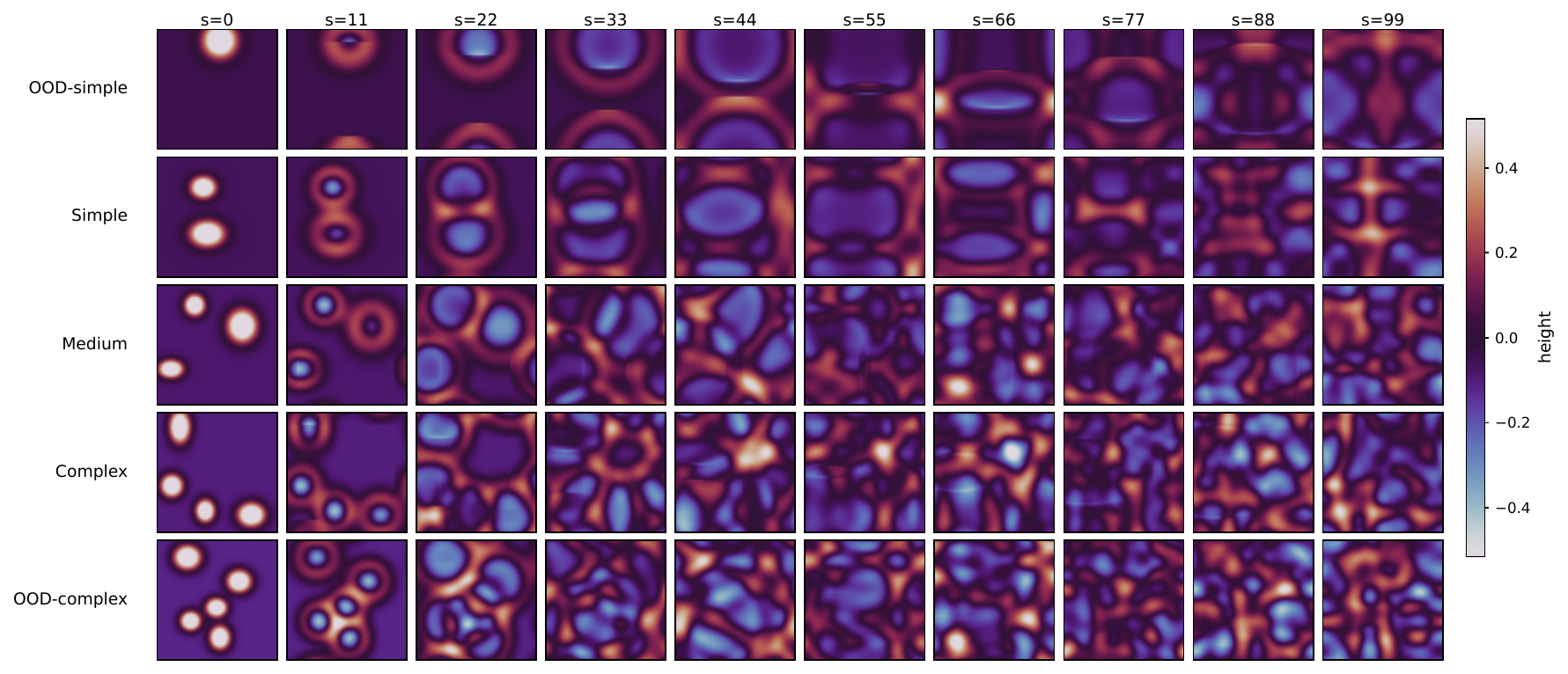}
    \caption{Wave-equation examples. Rows are IC families ordered from OOD-simple to OOD-complex; columns are uniformly sampled times.}
    \label{fig:pde-wave}
\end{figure}

The wave benchmark uses the second-order equation
\begin{equation}
    \partial_{tt} h = c^2 \Delta h,
\end{equation}
represented internally as the first-order system \(\partial_t h=v\), \(\partial_t v=c^2\Delta h\). We store both height \(h\) and velocity \(v\), and visualize \(h\) in \autoref{fig:pde-wave}. The domain extent is \(L=1.0\), the resolution is \(64\times64\), the wave speed is \(c=1.0\), and saved frames use \(\Delta t=0.01\) with one internal step per frame. The height initial condition is generated from Gaussian blobs, while the initial velocity amplitude is zero. Blob centers are sampled in the coordinate range \([0.1,0.9]\), variances are in \([0.002,0.008]\), and the field is normalized to zero mean and unit max absolute value. IC complexity is controlled only by the number of blobs: 1, 2, 3, 4, and 5 from OOD-simple to OOD-complex.

\subsection{Fisher--KPP}\label{app:data-fisher-kpp}

\begin{figure}[h]
    \centering
    \includegraphics[width=\linewidth]{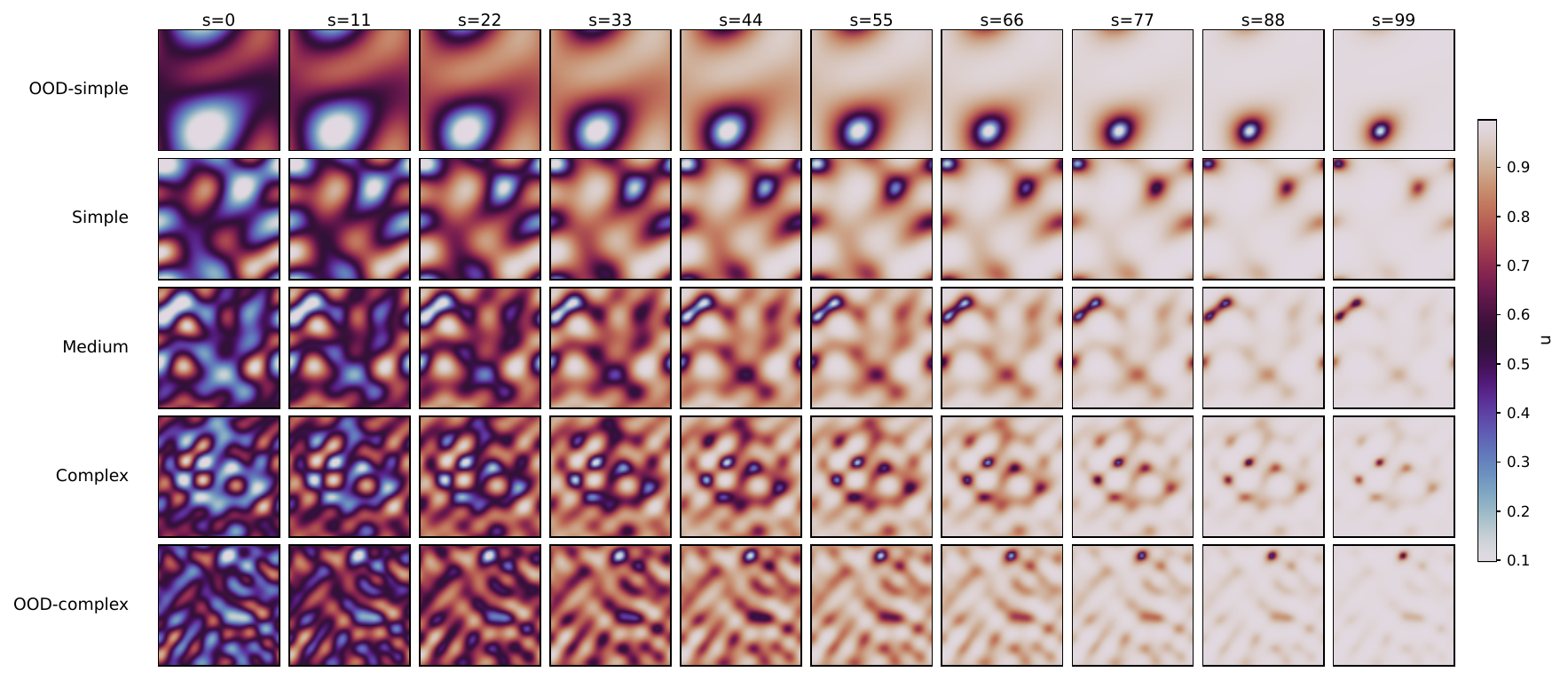}
    \caption{Fisher--KPP examples. Rows are IC families ordered from OOD-simple to OOD-complex; columns are uniformly sampled times.}
    \label{fig:pde-fisher-kpp}
\end{figure}

The Fisher--KPP equation is
\begin{equation}
    \partial_t u = D\Delta u + \rho u(1-u),
\end{equation}
with a single population channel \(u\). We use \(L=4.0\), \(64\times64\) resolution, solver step \(\Delta t=0.001\), two internal steps per stored frame, diffusivity \(D=0.005\), and reactivity \(\rho=20.0\). The saved frame spacing is therefore \(0.002\). Initial conditions are random truncated Fourier fields clamped and rescaled to the valid range \([0,1]\). Complexity is controlled by the Fourier cutoff: 1, 2, 3, 4, and 5 for OOD-simple, simple, medium, complex, and OOD-complex respectively. Representative trajectories are shown in \autoref{fig:pde-fisher-kpp}.

\subsection{Burgers}\label{app:data-burgers}

\begin{figure}[h]
    \centering
    \includegraphics[width=\linewidth]{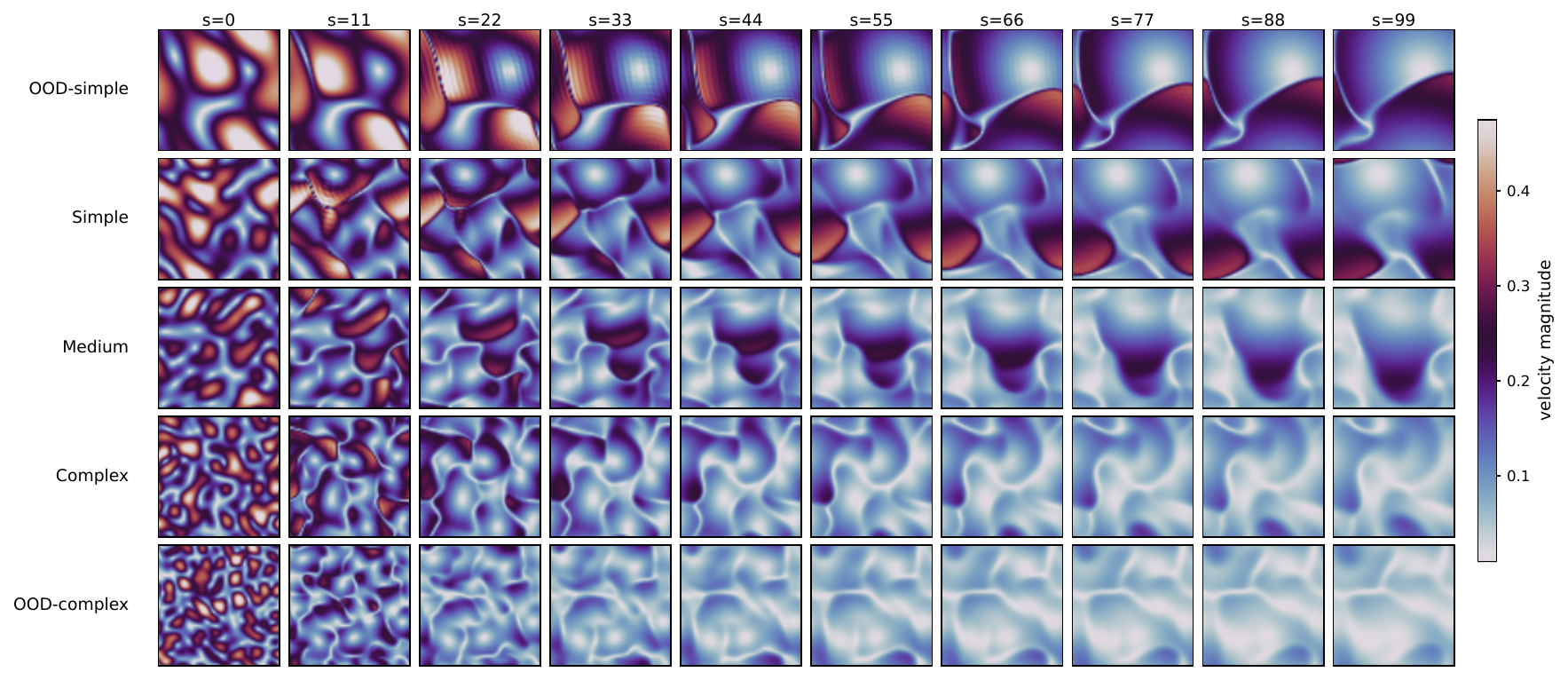}
    \caption{Burgers examples. Rows are IC families ordered from OOD-simple to OOD-complex; columns are uniformly sampled times.}
    \label{fig:pde-burgers}
\end{figure}

The two-dimensional Burgers benchmark stores the vector velocity field \(u=(u_x,u_y)\) and solves
\begin{equation}
    \partial_t u + \frac{b}{2}\nabla\cdot(u\otimes u) = \nu \Delta u .
\end{equation}
We use periodic boundary conditions, \(L=3.0\), \(64\times64\) resolution, solver step \(\Delta t=0.02\), and two internal steps per stored frame. Parameters are diffusivity \(\nu=0.01\), convection scale \(b=2.0\), amplitude 0.5, ETDRK order 2, and Orszag \(2/3\) dealiasing. \autoref{fig:pde-burgers} visualizes the velocity magnitude \(\sqrt{u_x^2+u_y^2}\). Initial conditions are random truncated Fourier fields; complexity is controlled by cutoff values 1, 2, 3, 4, and 5 from OOD-simple to OOD-complex.

\subsection{Swift--Hohenberg}\label{app:data-swift-hohenberg}

\begin{figure}[h]
    \centering
    \includegraphics[width=\linewidth]{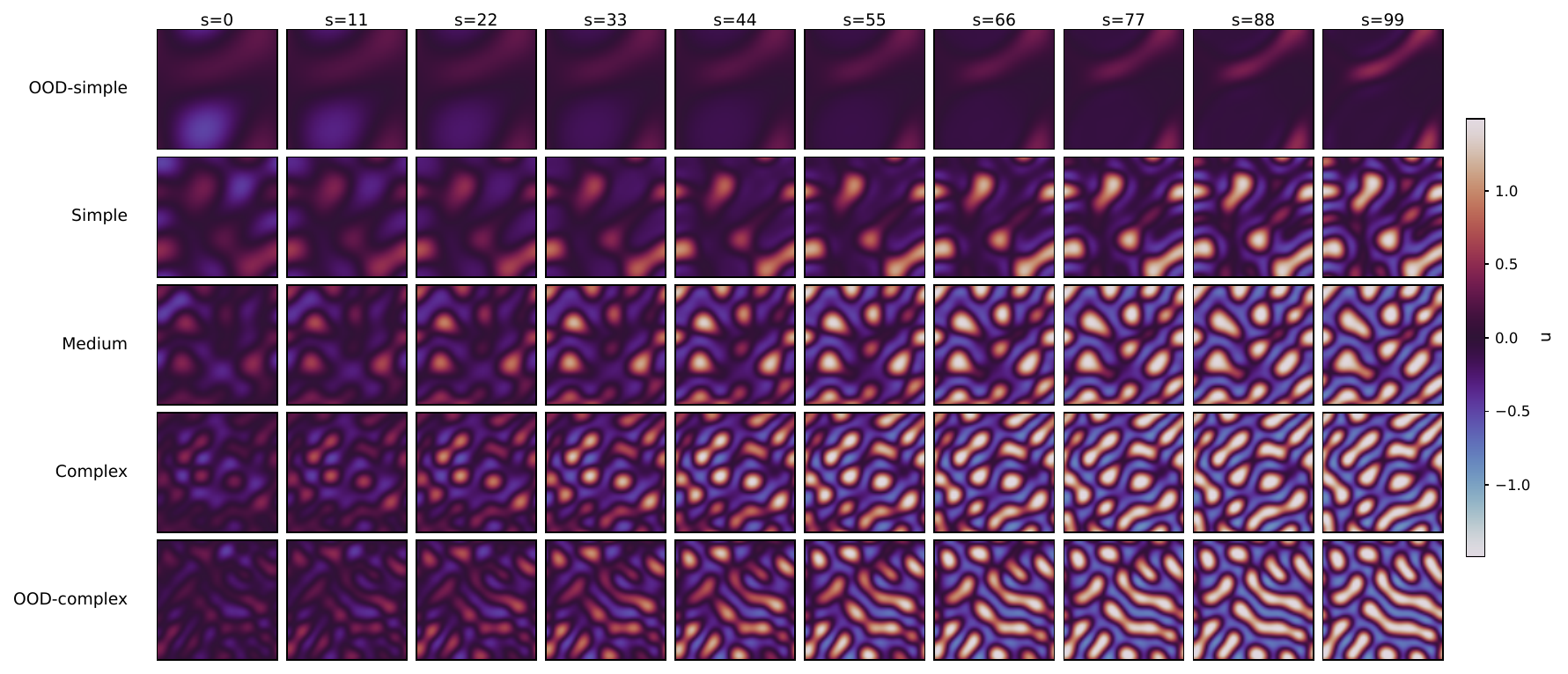}
    \caption{Swift--Hohenberg examples. Rows are IC families ordered from OOD-simple to OOD-complex; columns are uniformly sampled times.}
    \label{fig:pde-swift-hohenberg}
\end{figure}

The Swift--Hohenberg benchmark follows the Exponax reaction--diffusion form
\begin{equation}
    \partial_t u = r u - (q_c + \Delta)^2u + g(u),
    \qquad g(u)=u^2-u^3 .
\end{equation}
It is a single-channel pattern-formation system with periodic boundary conditions. We use \(L=30.0\), \(64\times64\) resolution, saved-frame spacing \(\Delta t=0.05\), reactivity \(r=0.7\), critical number \(q_c=1.0\), amplitude 0.5, ETDRK order 2, and dealiasing fraction \(1/2\). Initial conditions are random truncated Fourier fields with cutoff values 1, 2, 3, 4, and 5 from OOD-simple to OOD-complex. \autoref{fig:pde-swift-hohenberg} shows example temporal evolution.

\subsection{Decay}\label{app:data-decay}

\begin{figure}[h]
    \centering
    \includegraphics[width=\linewidth]{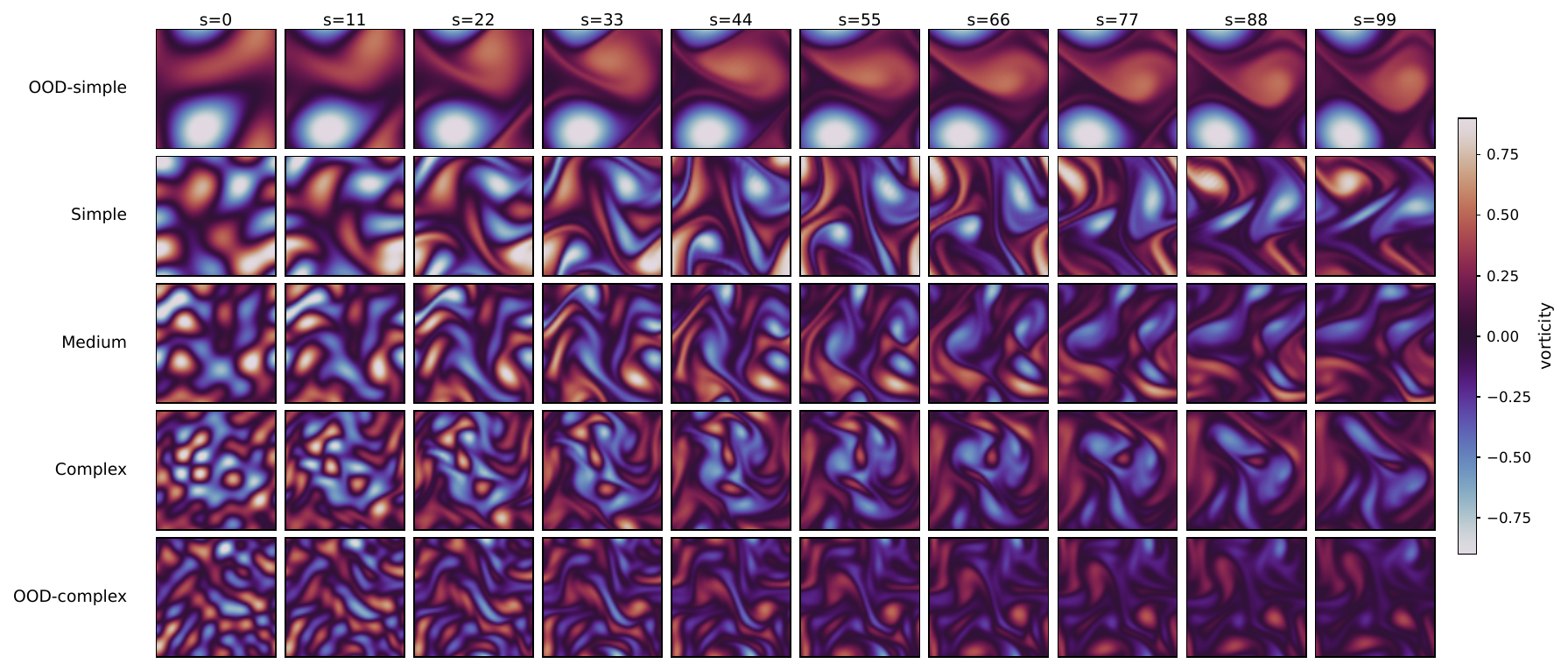}
    \caption{Navier--Stokes decay examples. Rows are IC families ordered from OOD-simple to OOD-complex; columns are uniformly sampled times.}
    \label{fig:pde-decay}
\end{figure}

The decay benchmark uses the two-dimensional incompressible Navier--Stokes equations in streamfunction--vorticity form. The stored scalar state is vorticity \(\omega\), governed by
\begin{equation}
    \partial_t \omega + b\,u\cdot\nabla \omega = \lambda \omega + \nu \Delta \omega,
    \qquad u = \nabla^\perp \psi, \quad \Delta \psi = \omega .
\end{equation}
We use \(L=5.0\), \(64\times64\) resolution, solver step \(\Delta t=0.06\), five internal steps per stored frame, diffusivity \(\nu=9\times10^{-4}\), convection scale \(b=1.0\), drag \(\lambda=0.0\), and amplitude 1.0. Initial conditions are random truncated Fourier vorticity fields with cutoff values 1, 2, 3, 4, and 5 from OOD-simple to OOD-complex. \autoref{fig:pde-decay} visualizes the vorticity field.

\subsection{Kolmogorov Flow}\label{app:data-kolmogorov}

\begin{figure}[h]
    \centering
    \includegraphics[width=\linewidth]{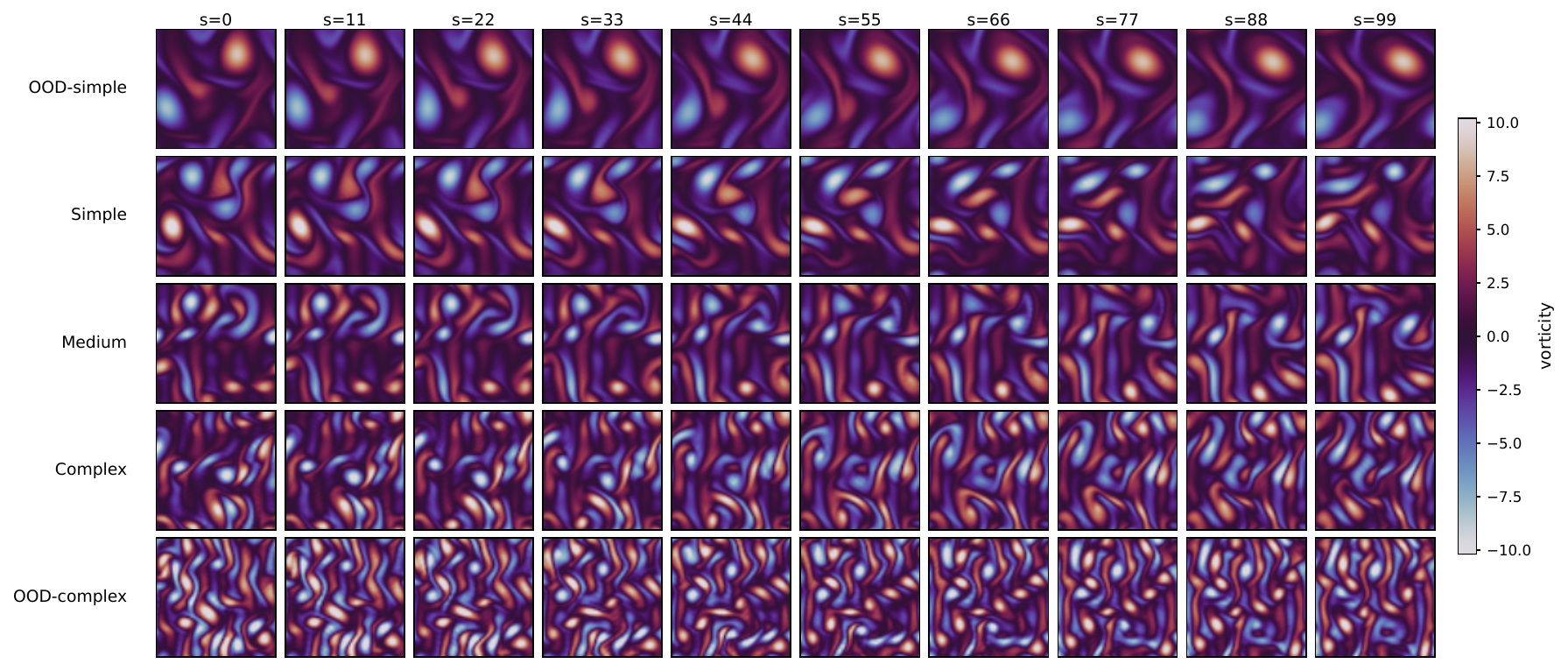}
    \caption{Kolmogorov-flow examples. Rows are IC families ordered from OOD-simple to OOD-complex; columns are uniformly sampled times.}
    \label{fig:pde-kolmogorov}
\end{figure}

Kolmogorov flow is also generated in streamfunction--vorticity form, but with a sinusoidal forcing term:
\begin{equation}
    \partial_t \omega + b\,u\cdot\nabla\omega = \lambda \omega + \nu\Delta\omega + f_m,
    \qquad u=\nabla^\perp\psi, \quad \Delta\psi=\omega .
\end{equation}
Here \(f_m\) is the curl of a Kolmogorov forcing at injection mode \(m\). We use \(L=6.0\), \(64\times64\) resolution, solver step \(\Delta t=0.01\), two internal steps per stored frame, diffusivity \(\nu=0.01\), convection scale \(b=1.2\), drag \(\lambda=-0.1\), injection scale 1.2, and amplitude 1.0. Each trajectory is simulated for 600 frames and the first 500 frames are discarded, so the stored 100 frames start after the initial transient. Complexity is controlled by the injection mode: \(m=2,3,4,5,6\) for OOD-simple, simple, medium, complex, and OOD-complex respectively. \autoref{fig:pde-kolmogorov} shows the retained vorticity dynamics.

\subsection{Kuramoto--Sivashinsky}\label{app:data-kuramoto-sivashinsky}

\begin{figure}[h]
    \centering
    \includegraphics[width=\linewidth]{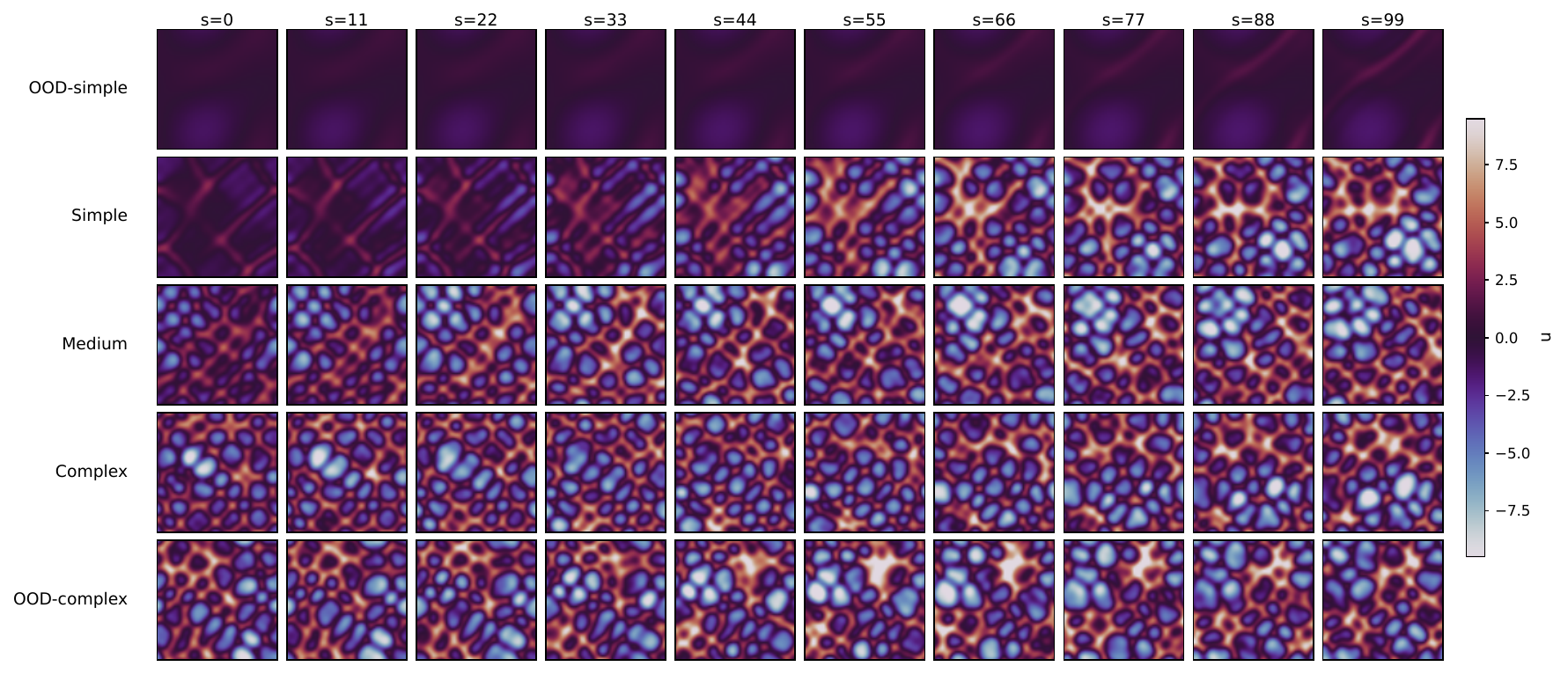}
    \caption{Kuramoto--Sivashinsky examples. Rows are IC families ordered from OOD-simple to OOD-complex; columns are uniformly sampled times.}
    \label{fig:pde-kuramoto-sivashinsky}
\end{figure}

We use the non-conservative, combustion-format Kuramoto--Sivashinsky equation implemented in Exponax:
\begin{equation}
    \partial_t u + \frac{b_2}{2}\lVert\nabla u\rVert_2^2 + \psi_1\Delta u + \psi_2\mathcal{D}_4 u = 0,
\end{equation}
where \(\mathcal{D}_4\) denotes the Exponax fourth-order derivative operator. The benchmark uses one scalar channel, \(L=42.0\), \(64\times64\) resolution, solver step \(\Delta t=0.1\), one internal step per stored frame, and scales \(b_2=\psi_1=\psi_2=1.0\). Each trajectory is generated for 200 frames and the first 100 are discarded before saving the final 100 frames. Initial conditions are random truncated Fourier fields with cutoff values 1, 2, 3, 4, and 5 from OOD-simple to OOD-complex. \autoref{fig:pde-kuramoto-sivashinsky} shows examples after transient trimming.

\section{Appendix: Supplementary Results}\label{app:diag}


\begin{figure}[p]
    \centering
    \includegraphics[width=\linewidth]{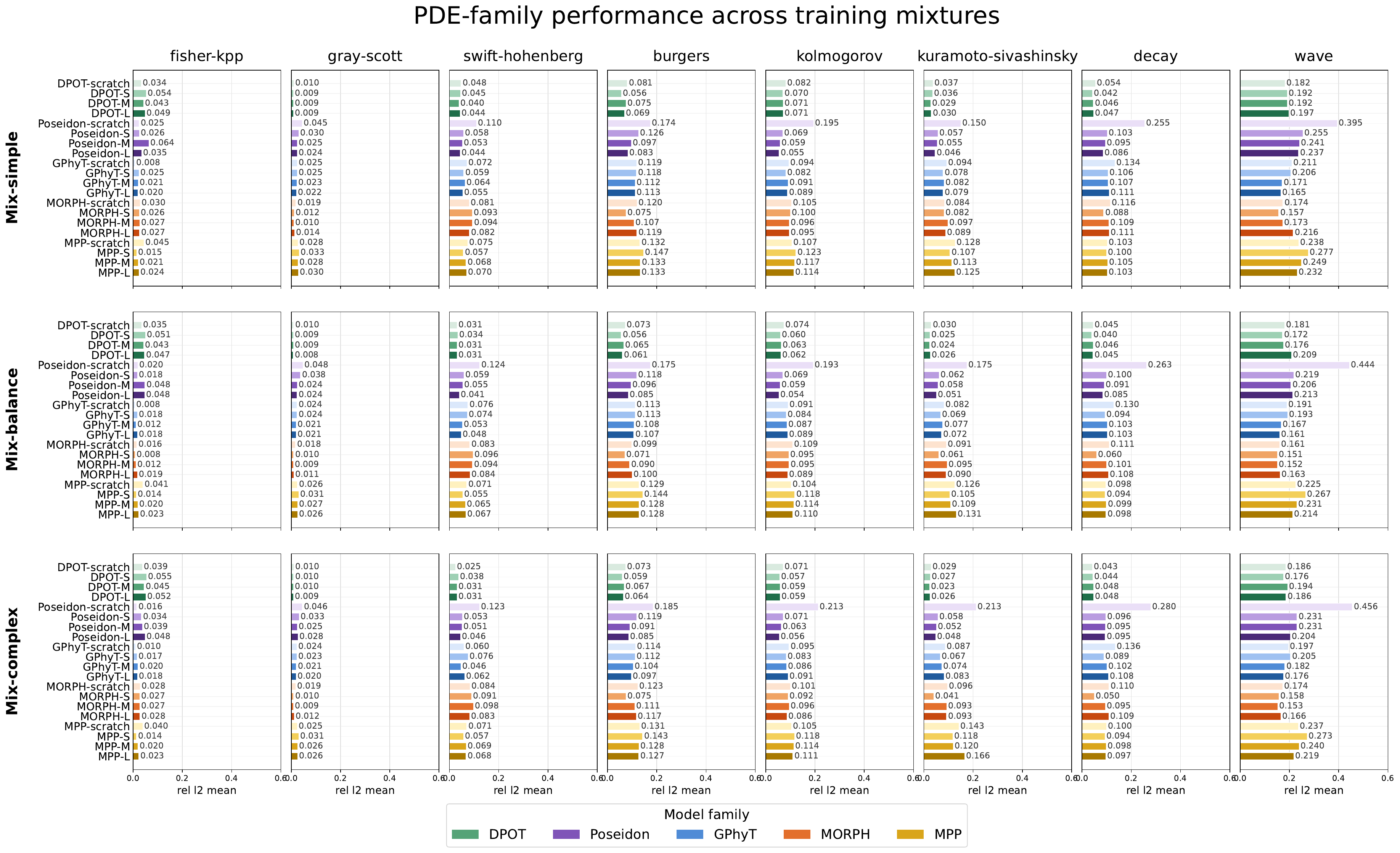}
    \caption{\textbf{PDE-family raw error across training mixtures at the first predicted frame.}
    Rows correspond to training mixtures and columns correspond to PDE families.
    Within each panel, bars show the raw relative $L_2$ error of each model
    variant, averaged over all 25 test regimes. Colors group variants by model
    family.}
    \label{fig:app_training_mix_pde_bar_id01}
\end{figure}

\begin{figure}[p]
    \centering
    \includegraphics[width=\linewidth]{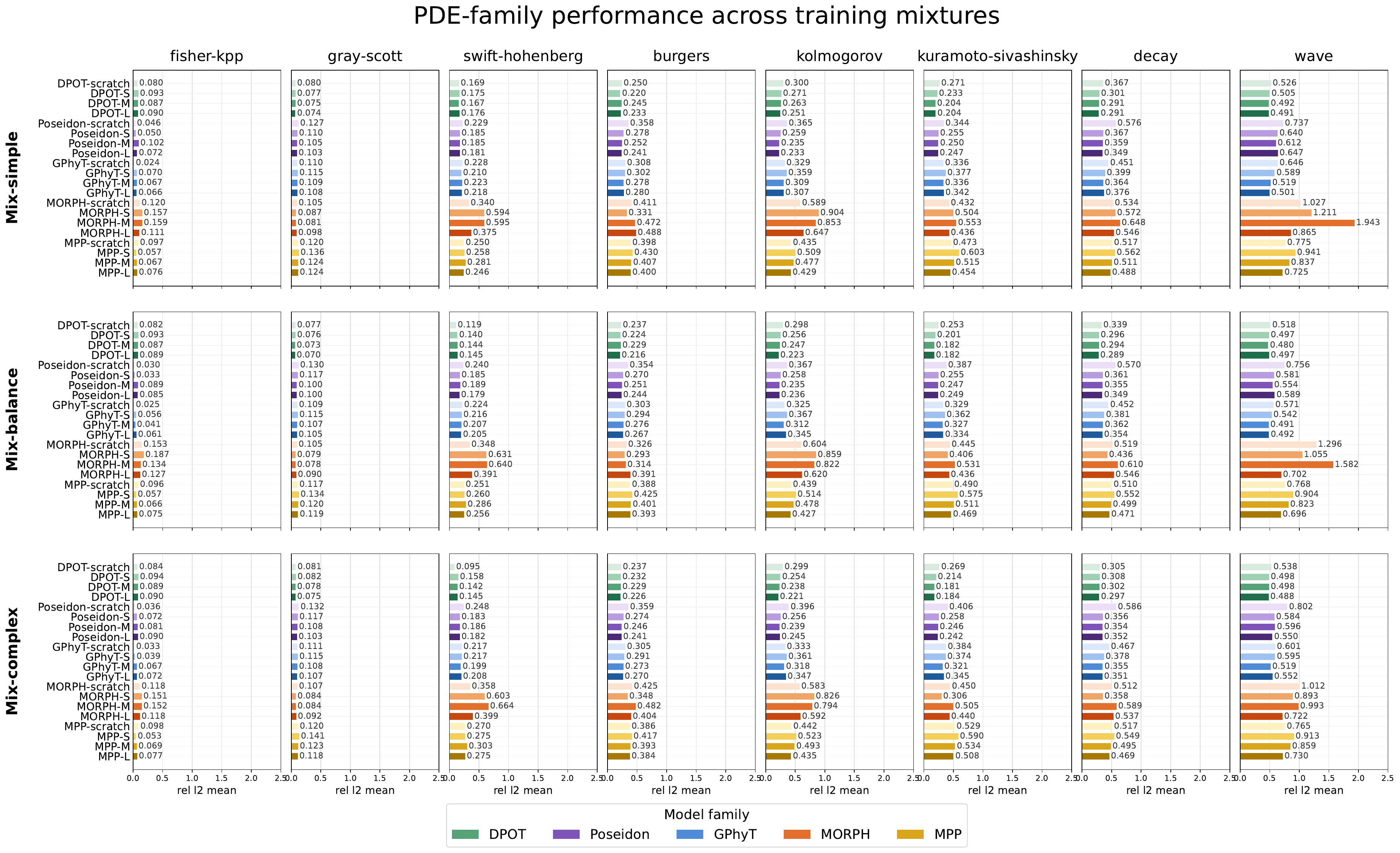}
    \caption{\textbf{PDE-family raw error across training mixtures at the 10-step horizon.}
    Rows correspond to training mixtures and columns correspond to PDE families.
    Within each panel, bars show the raw relative $L_2$ error of each model
    variant, averaged over all 25 test regimes and the 10 in-horizon predicted
    frames. Colors group variants by model family.}
    \label{fig:app_training_mix_pde_bar_id10}
\end{figure}

\begin{figure}[p]
    \centering
    \includegraphics[width=\linewidth]{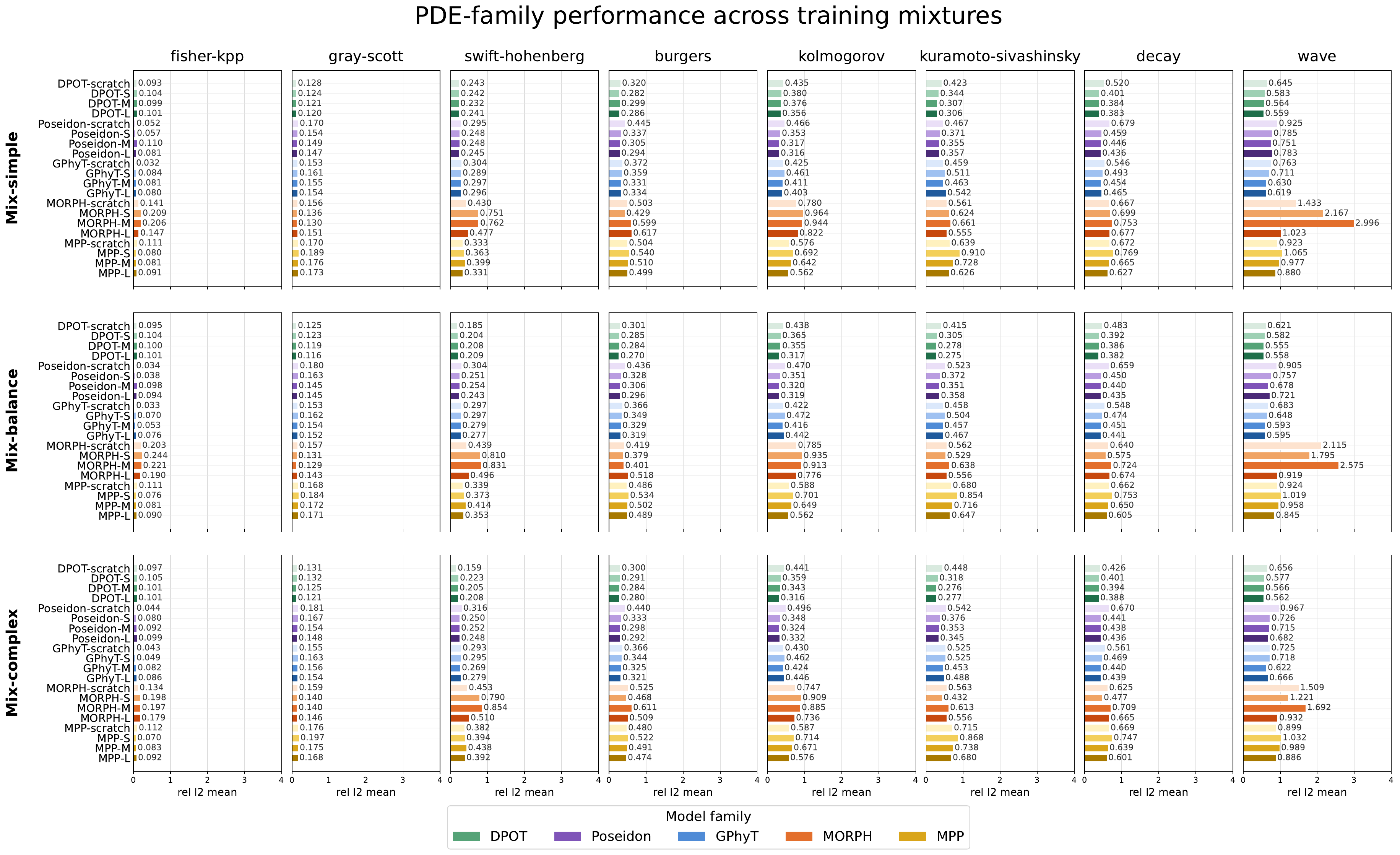}
    \caption{\textbf{PDE-family raw error across training mixtures over the full 15-frame rollout.}
    Rows correspond to training mixtures and columns correspond to PDE families.
    Within each panel, bars show the raw relative $L_2$ error of each model
    variant, averaged over all 25 test regimes and all 15 predicted frames.
    Colors group variants by model family.}
    \label{fig:app_training_mix_pde_bar_overall15}
\end{figure}

\begin{figure}[h]
    \centering
    \includegraphics[width=\linewidth]{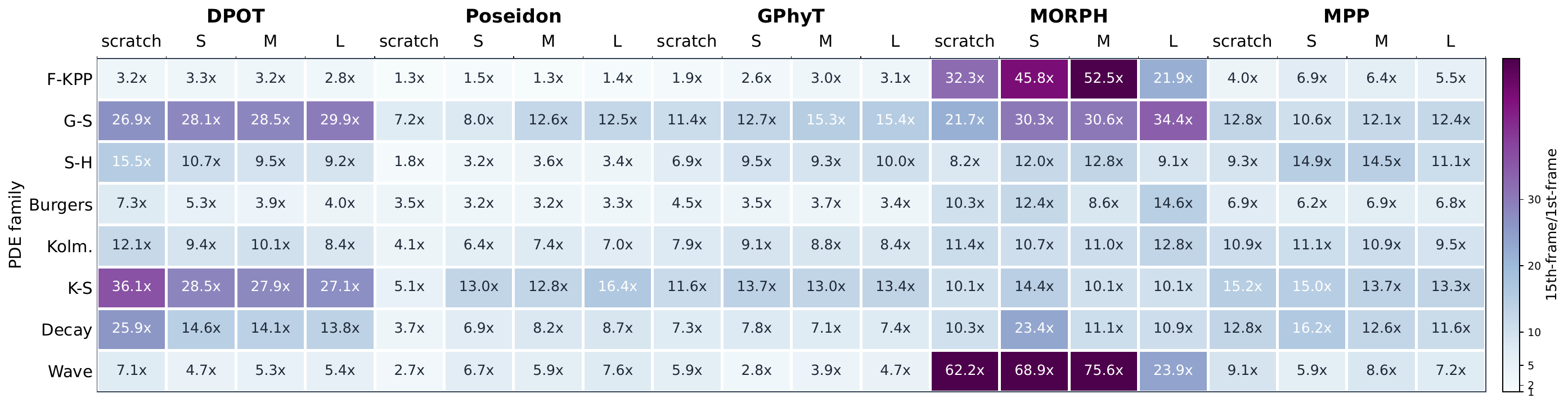}
    \caption{\textbf{15th-frame rollout amplification by model variant and PDE family.}
    Each cell reports
    $E_{\mathrm{15th\text{-}frame}}/E_{\mathrm{1st\text{-}frame}}$, where
    $E_{\mathrm{15th\text{-}frame}}$ is the error at prediction frame 15 and
    $E_{\mathrm{1st\text{-}frame}}$ is the first predicted-frame error. This
    figure is the rollout-horizon counterpart
    of \autoref{fig:rq4_amplification_strip}.}
    \label{fig:app_post_horizon_amplification}
\end{figure}

\begin{figure}[h]
    \centering
    \includegraphics[width=\linewidth]{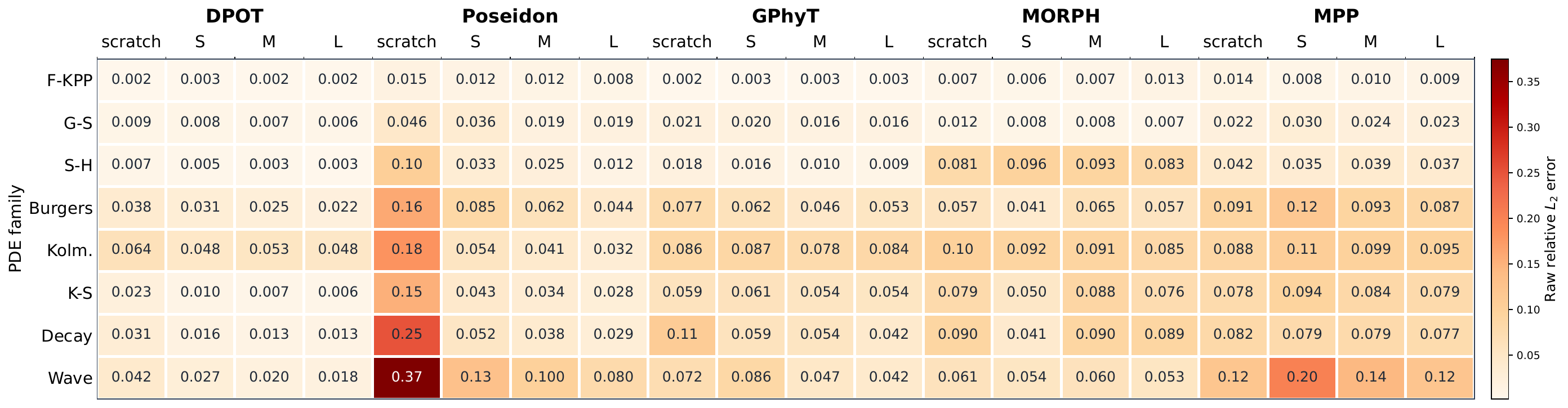}
    \caption{\textbf{Raw first-frame error by model variant and PDE family.}
    Each cell reports the raw relative $L_2$ error at prediction frame 1,
    averaged over the three train-seen cells under Mix-balance. Rows correspond
    to PDE families and columns correspond to model variants. The OrRd color
    range is normalized using the minimum and maximum values within this
    figure.}
    \label{fig:app_frame1_raw_error}
\end{figure}

\begin{figure}[h]
    \centering
    \includegraphics[width=\linewidth]{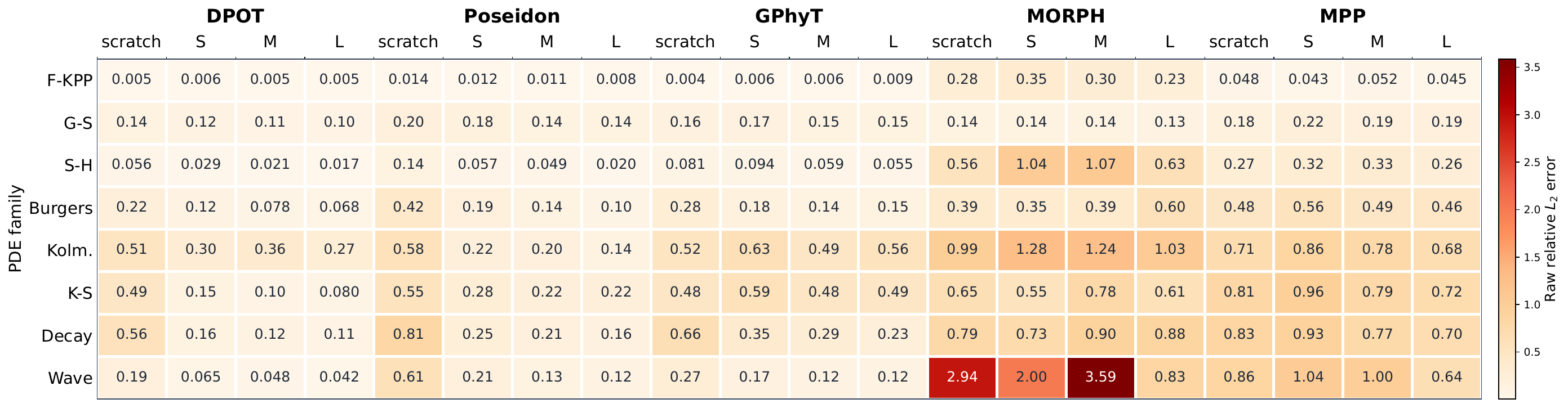}
    \caption{\textbf{Raw 10th-frame error by model variant and PDE family.}
    Each cell reports the raw relative $L_2$ error at prediction frame 10,
    averaged over the three train-seen cells under Mix-balance. Rows correspond
    to PDE families and columns correspond to model variants. The OrRd color
    range is normalized using the minimum and maximum values within this
    figure.}
    \label{fig:app_frame10_raw_error}
\end{figure}

\begin{figure}[h]
    \centering
    \includegraphics[width=\linewidth]{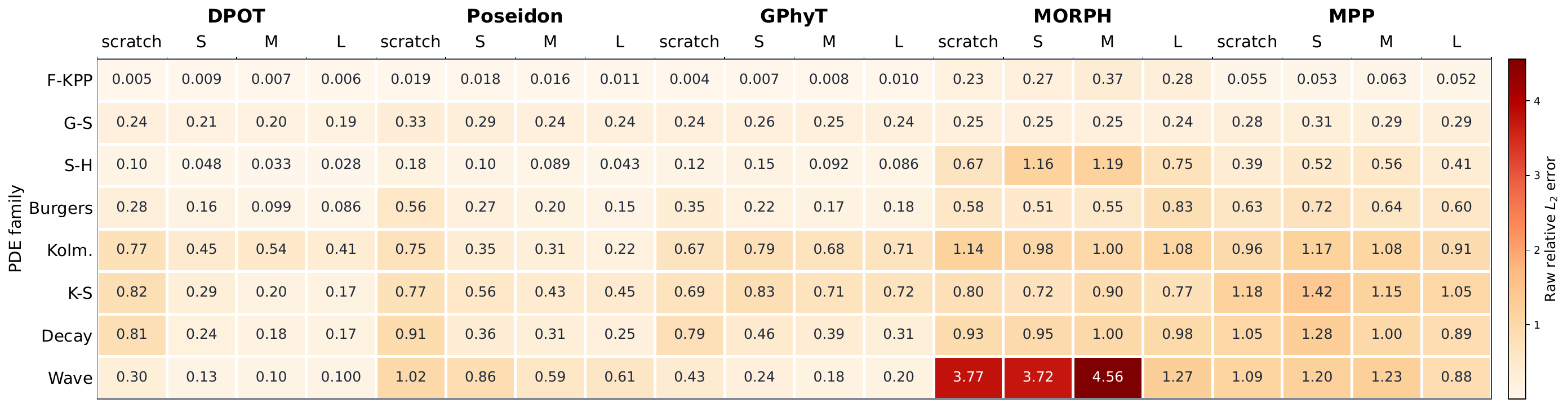}
    \caption{\textbf{Raw 15th-frame error by model variant and PDE family.}
    Each cell reports the raw relative $L_2$ error at prediction frame 15,
    averaged over the three train-seen cells under Mix-balance. Rows correspond
    to PDE families and columns correspond to model variants. The OrRd color
    range is normalized using the minimum and maximum values within this
    figure.}
    \label{fig:app_frame15_raw_error}
\end{figure}

\begin{figure}[h]
    \centering
    \includegraphics[width=\linewidth]{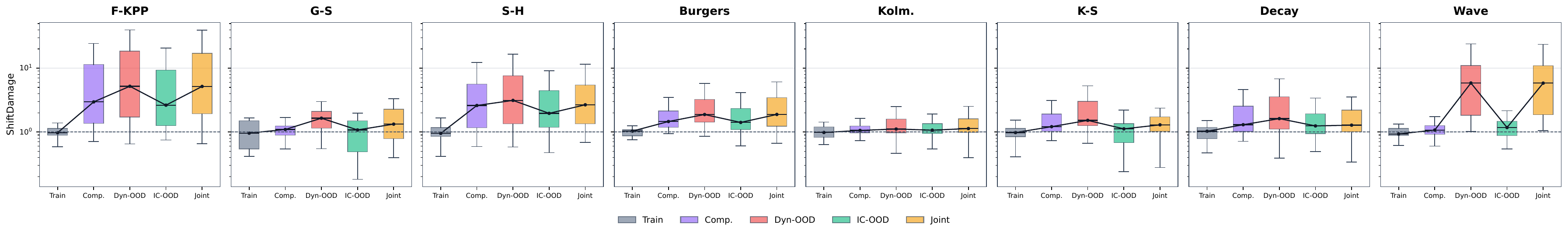}
    \caption{\textbf{ShiftDamage by shift group.}
    The same 10-step grouping shown separately for each PDE family. The dashed line
    marks no degradation relative to the train-seen baseline. Dynamic-OOD and
    Joint-OOD have the heaviest tails, indicating that dynamic-scale
    extrapolation produces larger relative robustness gaps than IC-only shifts
    on average, although the severity is PDE-dependent.}
    \label{fig:app_rq3_shift_groups}
\end{figure}

\begin{figure}[h]
    \centering
    \includegraphics[width=\linewidth]{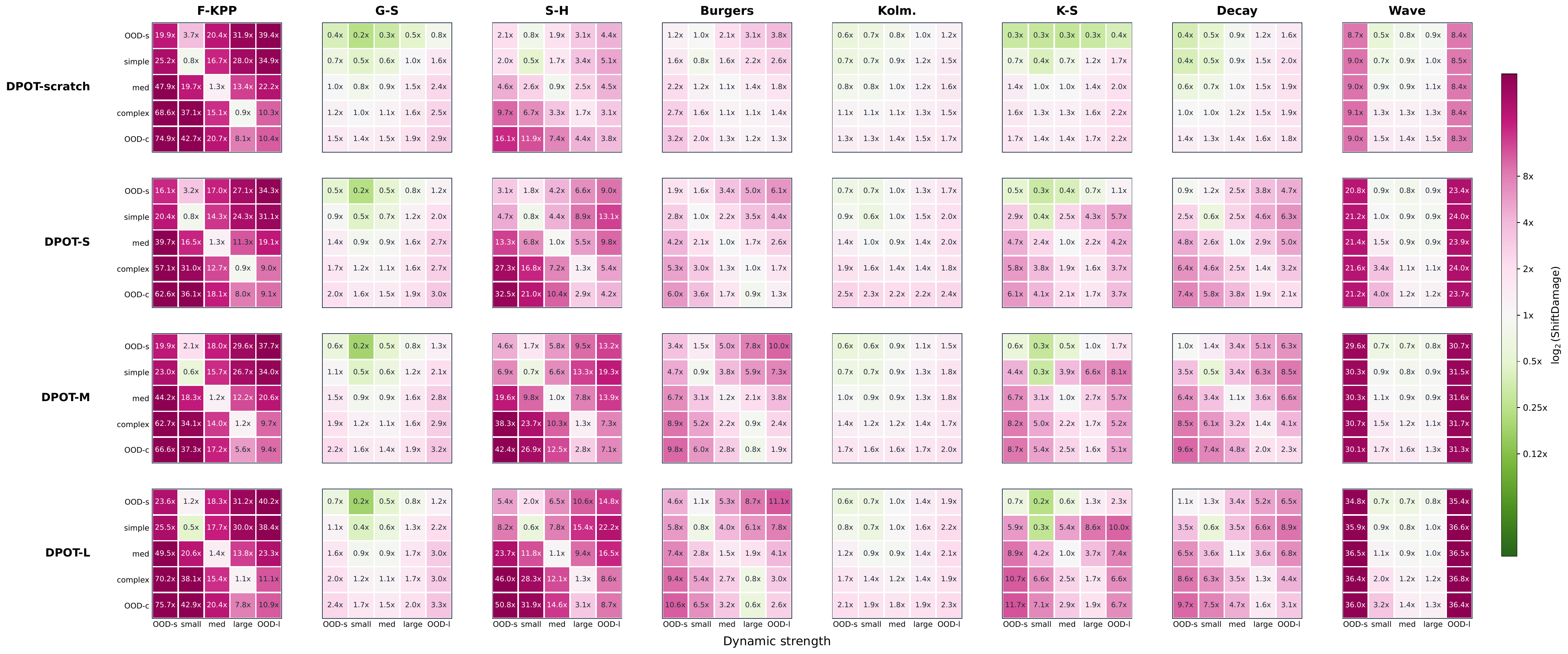}
    \caption{\textbf{DPOT variant-level $5\times5$ ShiftDamage grids.}
    Rows correspond to scratch/S/M/L variants, and columns correspond to PDE
    families. Each mini-grid shows 10-step ShiftDamage over dynamic strength and
    initial-condition complexity under Mix-balance. Colors use
    $\log_2(\mathrm{ShiftDamage})$: $1\times$ maps to zero/white, values below
    $1\times$ are green, and values above $1\times$ are red.}
    \label{fig:app_rq3_dpot_variant_5x5}
\end{figure}

\begin{figure}[h]
    \centering
    \includegraphics[width=\linewidth]{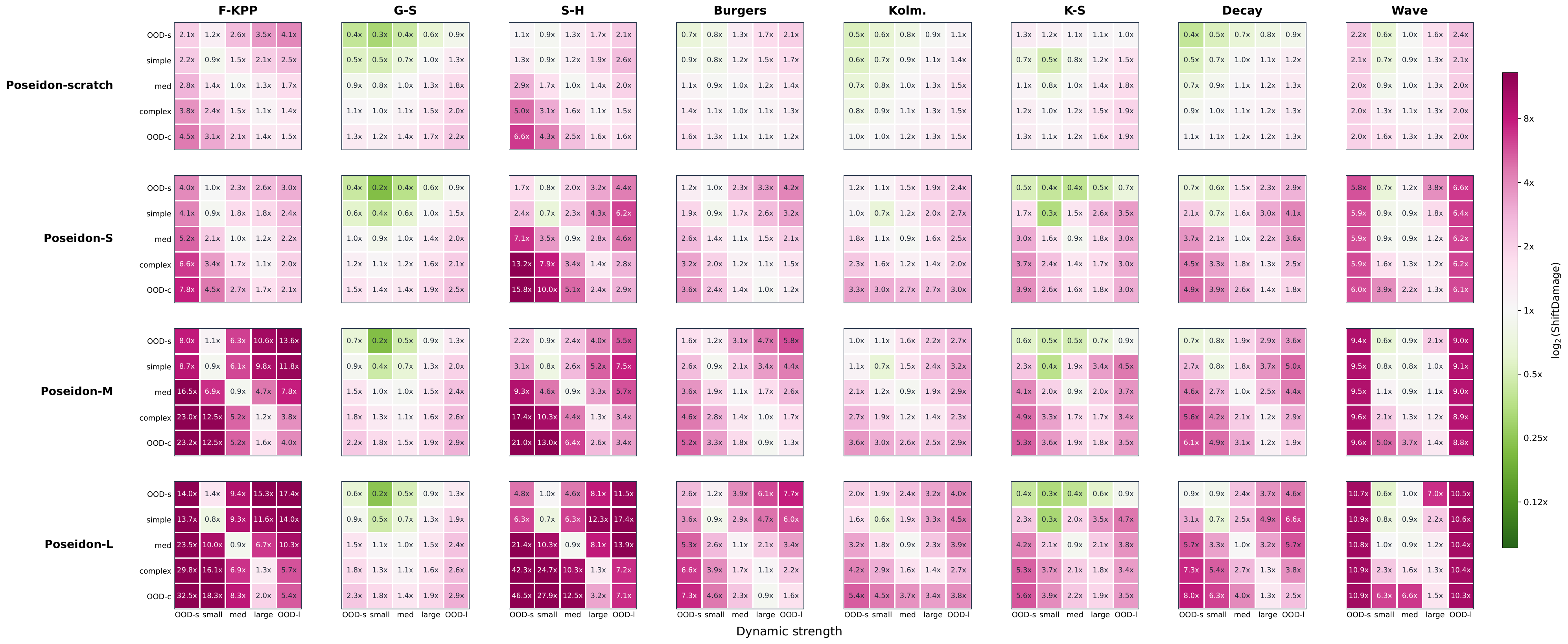}
    \caption{\textbf{Poseidon variant-level $5\times5$ ShiftDamage grids.}
    Rows correspond to scratch/S/M/L variants, and columns correspond to PDE
    families. Each mini-grid shows 10-step ShiftDamage over dynamic strength and
    initial-condition complexity under Mix-balance. Colors use
    $\log_2(\mathrm{ShiftDamage})$: $1\times$ maps to zero/white, values below
    $1\times$ are green, and values above $1\times$ are red.}
    \label{fig:app_rq3_poseidon_variant_5x5}
\end{figure}

\begin{figure}[h]
    \centering
    \includegraphics[width=\linewidth]{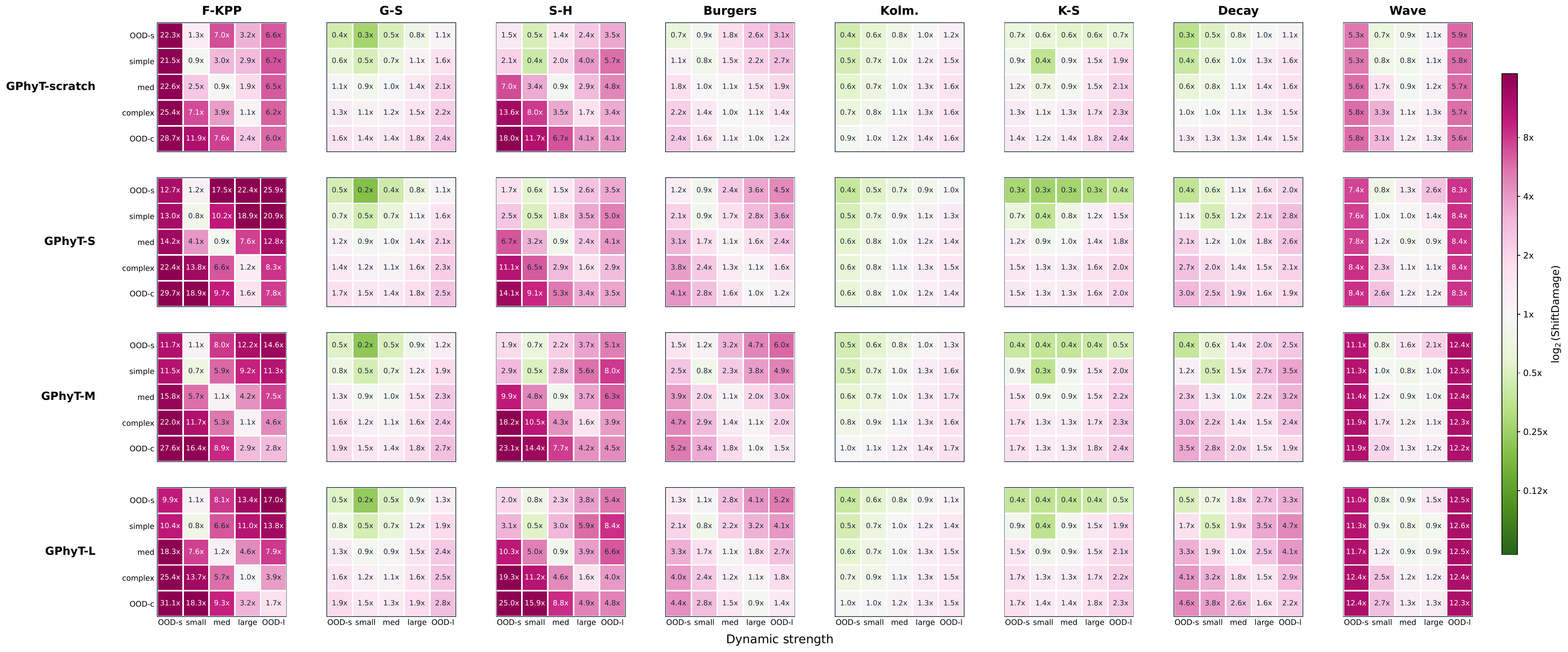}
    \caption{\textbf{GPhyT variant-level $5\times5$ ShiftDamage grids.}
    Rows correspond to scratch/S/M/L variants, and columns correspond to PDE
    families. Each mini-grid shows 10-step ShiftDamage over dynamic strength and
    initial-condition complexity under Mix-balance. Colors use
    $\log_2(\mathrm{ShiftDamage})$: $1\times$ maps to zero/white, values below
    $1\times$ are green, and values above $1\times$ are red.}
    \label{fig:app_rq3_gphyt_variant_5x5}
\end{figure}

\begin{figure}[h]
    \centering
    \includegraphics[width=\linewidth]{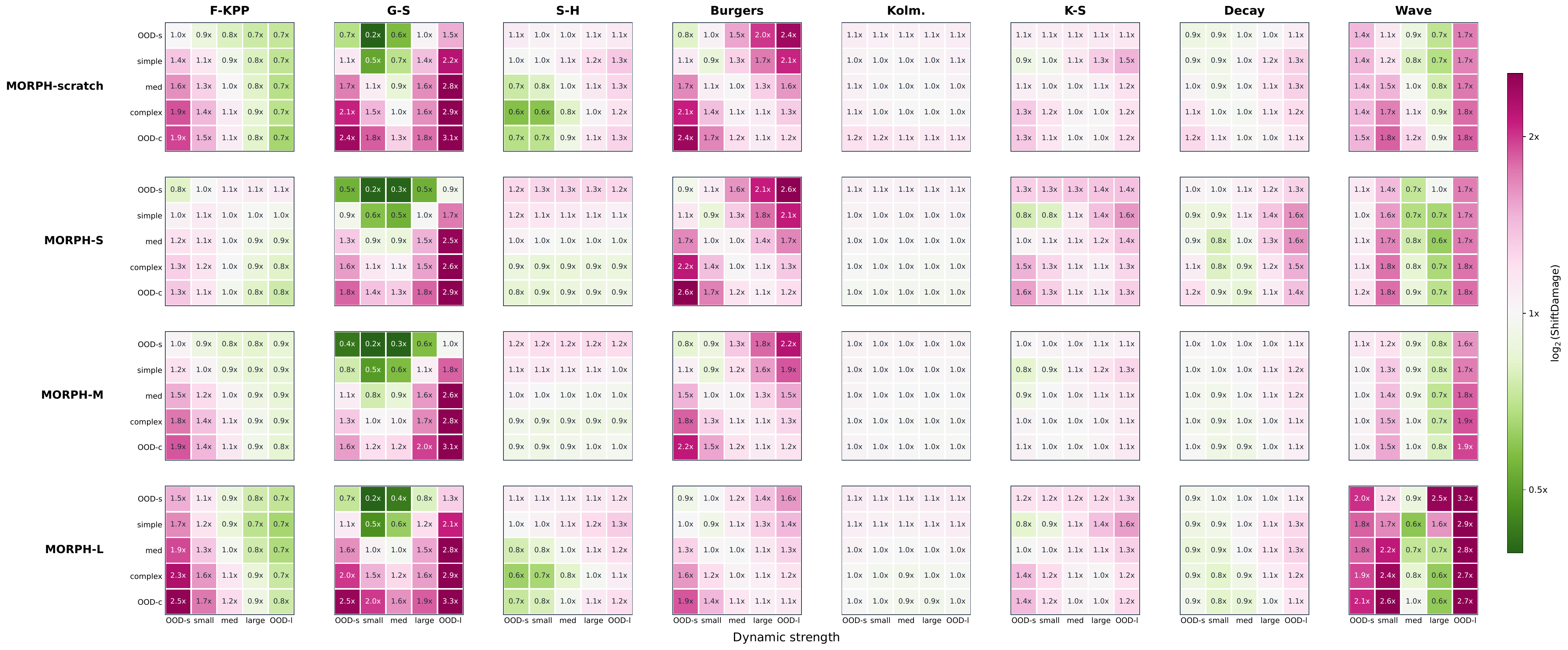}
    \caption{\textbf{MORPH variant-level $5\times5$ ShiftDamage grids.}
    Rows correspond to scratch/S/M/L variants, and columns correspond to PDE
    families. Each mini-grid shows 10-step ShiftDamage over dynamic strength and
    initial-condition complexity under Mix-balance. Colors use
    $\log_2(\mathrm{ShiftDamage})$: $1\times$ maps to zero/white, values below
    $1\times$ are green, and values above $1\times$ are red.}
    \label{fig:app_rq3_morph_variant_5x5}
\end{figure}

\begin{figure}[h]
    \centering
    \includegraphics[width=\linewidth]{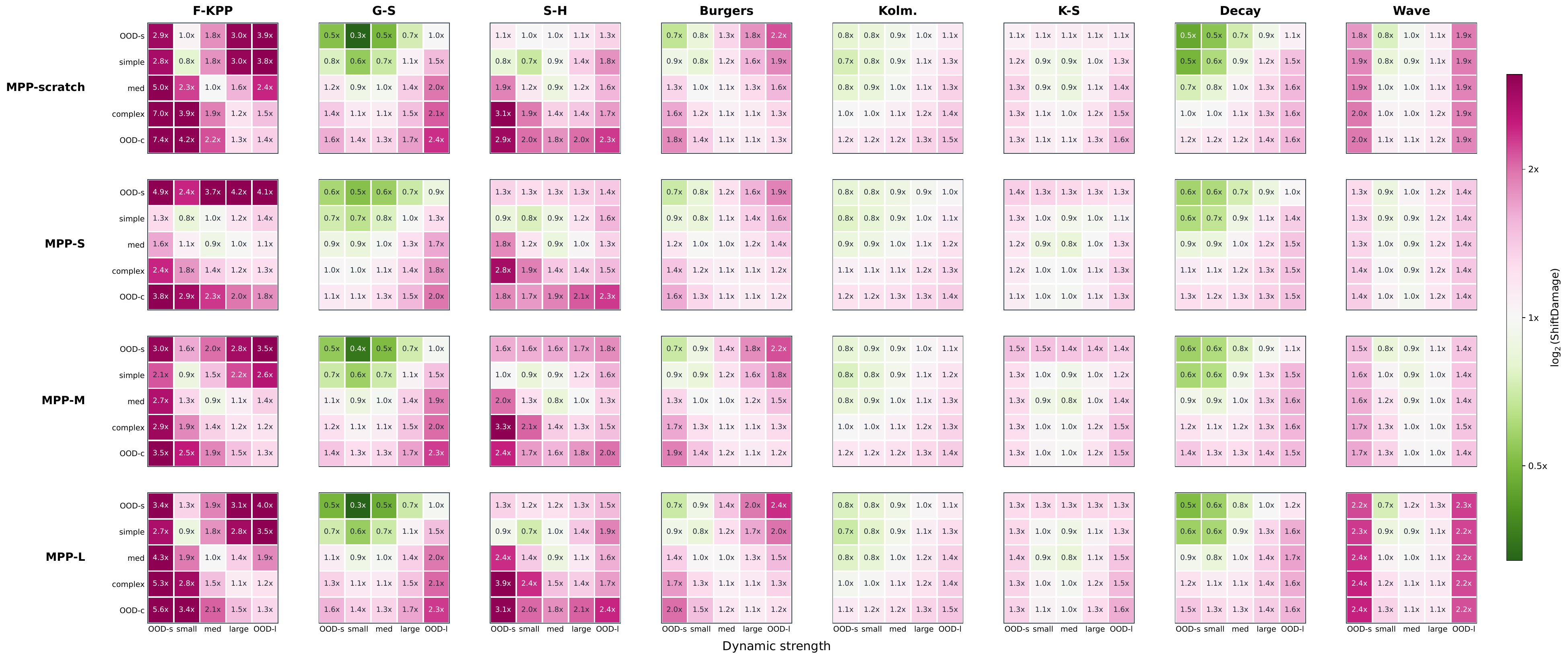}
    \caption{\textbf{MPP variant-level $5\times5$ ShiftDamage grids.}
    Rows correspond to scratch/S/M/L variants, and columns correspond to PDE
    families. Each mini-grid shows 10-step ShiftDamage over dynamic strength and
    initial-condition complexity under Mix-balance. Colors use
    $\log_2(\mathrm{ShiftDamage})$: $1\times$ maps to zero/white, values below
    $1\times$ are green, and values above $1\times$ are red.}
    \label{fig:app_rq3_mpp_variant_5x5}
\end{figure}

\end{document}